# Population Survey-Based Machine Learning Reveals Associations Between Psychosocial Factors and Chronic Kidney Disease

## Authors

Md. Atik Shams[1 9], David Eisenberg[5], Sumaiya Fatema[1], Asma Sultana[1], D. M Hasibul Islam[1], Junnatul Mawa[1], Anindita Datta[8], Nafiya Ahmed[2], Danastan Tasaouf Mridula[3], SK. Sazid Mahmud[4], Simon Bin Akter[3], Tanjila Helaly[1], Jorge Fresneda Fernandez[6], Humayera Islam[7], Tanmoy Sarkar Pias [10 11 *]

[1] Department of Computer Science and Engineering, University of Asia Pacific, Dhaka 1205, Bangladesh
[2] Department of Computer Science, BRAC University, Dhaka 1212, Bangladesh
[3] Department of Computer Science and Engineering, Northern University Bangladesh, Dhaka 1230, Bangladesh
[4] Institute of Biological Sciences, Rajshahi University, Rajshahi 6205, Bangladesh
[5] Department of Information Management and Business Analytics, Montclair State University, Feliciano School of Business, NJ, USA
[6] Martin Tuchman School of Management, New Jersey Institute of Technology, Newark, 07102, NJ, USA
[7]Institute for Population and Precision Health, University of Chicago, Department of Family Medicine, Illinois, USA
[8]Department of Fish and Wildlife Conservation, College of Natural Resources and Environment, Virginia Tech, Blacksburg, VA 24060, USA
[9]Department of Computer Science and Engineering, Notre Dame University Bangladesh, Dhaka 1000, Bangladesh
[10]Department of Computer Science, College of Engineering, Virginia Tech, Blacksburg, VA 24060, USA
[11]School of Medicine, Stanford University, Stanford, CA 94305, USA

* corresponding author (tspias@stanford.edu)

## Abstract

Chronic kidney disease (CKD) progresses silently and severely undermines quality of life, making early detection critical for improving patient outcomes. We present a two-part study that combines large-scale telehealth data with advanced machine learning to both classify self-reported CKD status and identify key drivers of disease. Using selected features from the Behavioral Risk Factor Surveillance System (BRFSS 2021: 438,693 samples; BRFSS 2019: 418,268 samples) and the National Health Interview Survey (NHIS 2021: 29,482 samples; NHIS 2020: 31,568 samples), we addressed missing data with nine state-of-the-art imputation methods and mitigated class imbalance via sampling strategies. Our customized stacked ensemble model achieved balanced accuracy of 72.56-76.12%, with corresponding AUROC scores of 79.59-82.29%. SHapley Additive exPlanations (SHAP) analysis, followed by clinical review, highlighted critical predictors, including regular medical check-ups, age, blood pressure, and indicators of mental health stress. These findings deliver a robust and interpretable framework for CKD risk stratification and provide actionable insights into its associated factors.

## Introduction

Chronic kidney disease (CKD) is one of the most life-threatening public health issues worldwide, which is anticipated to become the fifth major cause of death by 2050[1]. According to the Centers for Disease Control and Prevention (CDC), more than one in seven Americans (37 million people) are estimated to have CKD, and 40% of them are unaware of their condition[2]. This is an important diagnosis gap that shows a crucial limitation of databases, which are derived from hospitals. These datasets contain patients who are already under treatment. Thus, a major part of early stage cases remains unidentified[3]. On the other hand, cross-sectional surveys such as BRFSS and NHIS capture a broader population, allowing for the classification and risk stratification of individuals who share risk profiles with self-reported CKD patients[4]. But this type of self-reported dataset often contains noise and a high amount of missing values. To address this challenge, we propose a machine learning framework that combines different imputation techniques and stacking ensemble model to detect the early risk patterns of those missing data contain samples[5]. The long-term complications of CKD include traditional risk factors such as diabetes[6], obesity[7], hypertension[8], and adverse life events[9].

Early detection of CKD is vital to prevent these complications and patients' severe health outcomes related to the condition. Though a single test, such as serum creatinine and eGFR is comparatively not that much costly but overall deploying them in labs for mass screening leads to considerable operational and infrastructural costs[10,11]. The necessity for physical visits in hospitals, drawing blood and laboratory processing pose logistical challenges. Adopting such invasive methods is not very useful for a large population that doesn't have symptoms for early prevention. Recently, with the advancement of machine learning (ML), large-scale risk stratification and classification of CKD has become more cost-effective. Advanced ML and deep neural networks (DNNs) have demonstrated promising outcomes in several healthcare applications, including electroencephalograms[12], electrocardiograms[13], medical image analysis[14], drug discovery and development[15]. Indeed, in recent decades, health surveys have been extensively used in healthcare research[16,17], considering lifestyle habits, demographic information, individuals' past medical records, adverse childhood experiences, dietary habits, and several health indicators[18,19]. ML approaches achieve significant performance by leveraging the increasing availability of medical data, including clinical test reports and individual patients' medical records.

Nonetheless, researchers conducting research with health survey datasets typically exclude missing values[20,21], such as refusing to answer a particular question or replying they do not know/not sure[22,23]. Additionally, health survey data often show non-linear patterns, such as class imbalance, consequently machine learning models obtain inferior performance applying survey datasets for predicting disorders[24] and demonstrate the imbalance in the F1-score and sensitivity[25,26]. Consequently, advanced computational techniques are essential to discern important insights and mitigate class imbalance issues from these medical survey datasets[20,27]. Several studies[28–30] addressed imbalance in the F1-score by applying sampling techniques such as NearMiss, Tomek Links, Synthetic Minority Oversampling Technique (SMOTE), Random Undersampling (RUS), Random Oversampling (ROS), Adaptive Synthetic Sampling (ADASYN), Edited Nearest Neighbor (ENN), SMOTE-ENN, cost-sensitive learning, and threshold tuning. Also, traditional data imputation (mean, median, or mode) is often unreliable, especially for a large amount of missing values[31,32], which subsequently reduces accuracy in contrast with K-Nearest Neighbor (KNN), Autoencoders, Diffusion, and Matrix Factorization (MF) have been more effective in improving model accuracy[33,34]. Because of missing values and class imbalances in medical data, ultimately, all of these hinder the performance of the individual ML approaches[35]. Therefore, it is important to handle missing values carefully in health-related survey datasets. These datasets often suffer from significant class imbalance, with only a limited amount of data available for certain classes. Removing additional records due to missing values further reduces the presence of these underrepresented classes, worsening the imbalance and lowering model performance. To address this problem, advanced imputation methods and resampling can be used, making model predictions more accurate and fair.

A significant number of studies focus on ensemble methods in CKD prediction[36,37] to present a more generalized model instead of a single base ML model[38,39], as these models gained more notable performance in classifying CKD status[40,41]. For example, achieved an AUC of 0.883 using ensemble techniques to predict CKD progression among clinical patients[42]. Many of these studies rely on structured health surveys and clinical datasets, which often suffer from severe class imbalance, making prediction more challenging. Despite the rising popularity of ensemble approaches[38,42], there is still room to refine the existing ensemble methods to improve models' performance and real-world applicability. Hence, this study aims to deal with the existing limitations[43,44] by applying a stacking ensemble model for CKD detection.

Extensive research has explored the prediction of CKD using various ML models, however these approaches have notable limitations[45,46]. Prior studies have reported high performance using ML approaches on both hospital-based data[41,47] and the UCI CKD datasets[46,47]. Luo et al. (2024) achieved an AUC of 0.833 with 95.2% sensitivity and 71.4% specificity, while Ghosh et al. (2024) achieved an AUC of 0.9689 and 93.3% accuracy. Debal et al. (2022) achieved 99.8% accuracy for binary classification using RF and 82.56% for multi-class classification using XGBoost. Similarly, Dharmarathne et al. (2024) applied explainable ML to predict CKD and evaluated six ML models, with their best-performing model, XGBoost, achieving 97.5% accuracy, 98.7% precision, 97.4% recall, and an F1-score of 98, and Dey et al. (2022) reported 98% accuracy using the ETC. Chittora et al. (2021) applied seven ML classifiers and a deep neural network using feature selection techniques and only applied SMOTE to enhance predictive accuracy, achieving a highest accuracy of 99.6%. Despite these promising results, all of these studies relied on small and clean datasets that do not fully reflect real-world clinical data characteristics, including high class imbalance, substantial missing values, and the need for extensive preprocessing. For instance, did not address missing-value handling[46,48], relied solely on median imputation[44], applied only mean imputation[49] and used KNN imputation[47,50]. These basic approaches, rather than more advanced techniques, may not be effective in complex clinical contexts. Furthermore, the reliance on small datasets lacks robustness in handling large volumes of missing data, which is more representative of real-life scenarios. Bai et al. used five ML models (LR, NB, RF, DT, and KNN) to predict ESKD in CKD patients[51], imputing missing data via multiple imputation. The dataset contained 748 CKD patients from Peking University Third Hospital (2006–2008), with a follow-up period of 6.3 ± 2.3 years. However, limitations include its single-center dataset, lack of extensive feature selection analysis, and no application of explainability methods.

It is evident that experimenting with multiple imputation (MI) methods on the dataset rather than solely depending on one imputation technique is more effective, as different imputation algorithms capture different patterns in the missing data. For instance, simple methods like mean or median imputation can distort variance and weaken relationships, while advanced methods such as KNN, GAN, RF, and XGBoost work better. Studies have shown that using MI strategies improves robustness and reduces bias in healthcare datasets[52,53]. MI outperformed deletion and single imputation, highlighting its effectiveness for handling missing data in clinical datasets. They simulated missing values in a subset of cases and compared parameter estimates, with MI providing more accurate results than the simple method. It needs different approaches for managing missing data in large datasets such as NHANES and BRFSS, than approaches for small clinical datasets. Multiple Imputation by Chained Equations (MICE) has been considered as the standard for epidemiological data. It makes it possible to estimate error across imputed datasets[54]. But MICE often suffers from computational challenges with high-dimensional data and may poorly represent complex nonlinear correlations among features. Recent studies have started to use machine-learning-based approaches, such as Random Forest estimation (MissForest) and k-Nearest Neighbors (k-NN), which are more accurate with all kinds of datasets[55]. Recently, deep learning approaches such as Generative Adversarial Networks (GANs) and Denoising Autoencoders have become an effective way to handle large-scale imputation. These approaches have shown a better ability to rebuild data distributions without the complex distributional assumptions that traditional statistical methods need[56,57].

The BRFSS dataset has been widely used for predictive modeling of various health outcomes, including cognitive disability[58], multiple chronic diseases, COVID-19 mortality[59], and heart attack risk[60]. While these studies have contributed significantly to disease classification using BRFSS data, no one has specifically focused on CKD, and most importantly, the issue of many missing values in the BRFSS dataset features has not been addressed before. This limitation reduces the reliability of existing findings. Our study addressed this gap by applying multiple advanced imputation methods, along with resampling and explainable ML, to improve the robustness and interpretability of CKD classification. Similarly, the NHIS dataset has been widely used to study diverse health outcomes, including stroke classification, multiple chronic disease risk factors, breast cancer risk, and developmental disabilities associated with anxiety and depression[17,61]. These studies demonstrate the broad utility of NHIS for large-scale health research and predictive modeling. Previous studies have used national health datasets to examine the relationships of renal health, but prediction models for CKD classifications are scarce. Dharmarajan et al. (2017) utilized the BRFSS dataset to determine the knowledge of people in each state about CKD. They employed weighted logistic regression to demonstrate the differences in diagnosis[62]. Similarly, Salifu et al. (2014) used NHIS data from 2004 to 2006 to investigate the relationship between sleep length and self-reported CKD. They found that short sleep was a risk factor[63]. Peng et al. (2019) also used Association Rule Mining (ARM) on survey data to identify lifestyle trends associated with kidney disease[64]. These studies mostly focused on specific epidemiological links using standard statistical methods. Handling of missing data is important because deleting incomplete records reduces the effective sample size, worsens pre-existing class imbalance, and can compromise the reliability of predictive models. Simple deletion methods only result in unbiased estimates if the data is missing completely at random. However, data is often missing at random or not at random in complex health surveys. In such cases, deleting incomplete rows leads to significant bias in the features. This makes advanced imputation a necessity[65].

Previous studies have made noticeable advancements in predicting CKD. However, it often overlooks key methodological challenges. Most studies did not handle the combination of mixed data types, missing data, and severe class imbalance simultaneously. First, some studies depended on simple statistical models that were not very strong at predicting and did not clearly explain how they reached their results. Dharmarajan et al. used simple weighted logistic regression and Peng et al. focused on finding common data patterns using basic association rules[62,64]. Then, missing data handling remains a significant weakness. Chittora et al. and Luo et al. either ignored it fully or depended on a single method that cannot capture the complex patterns of clinical data[46,48]. Lastly, many studies faced the challenge of imbalanced medical data because they did not explore enough imputation techniques. They often use only one approach such as SMOTE[66,67], rather than comparing the performance of multiple imputation techniques. On the other hand, our study overcomes these limitations by applying ten advanced techniques for missing values and evaluating five different resampling methods. Finally, we designed Stacked Hybrid Architectures with SHAP to ensure the model's decisions are transparent.

This study incorporates a robust and effective stacking ensemble strategy for the cost-effective, large-scale classification of kidney ailment from the CDC's Behavioral Risk Factor Surveillance System (BRFSS) and from the National Health Interview Survey (NHIS), one of the largest ongoing national health surveys, considering demographic characteristics, health behaviors among United States adults, health insurance, disability, and eight Adverse Childhood Experiences (ACEs) features. This study employed nine data imputation techniques, including Forward-Backward, Mean, Matrix Factorization (MF), Generative Adversarial Networks (GAN), Diffusion, Autoencoder-5, Autoencoder-15, Drop-nan to handle missing values[68,69], and five sampling techniques, including No Sampling, Random Oversampling (ROS)[70,71], Random Undersampling (RUS)[70,71], Synthetic Minority Over-Sampling Technique (SMOTE)[72,73], and Adaptive Synthetic Sampling (AdaSyn)[74] after imputation to reduce class imbalance issues. Following, this study applied and analyzed eleven ML and DL techniques, including eXtreme Gradient Boosting (XGBoost), Linear Regression (LR), Adaptive Boosting (AdaBoost), Random Forest (RF), Light Gradient Boosting Machine (LightGBM), Decision Tree (DT), Gradient Boosting (GB), Artificial Neural Networks (ANN) with 1 and 2 hidden layers, ANN + Attention and Stacked Ensemble to systematically compare their predictive performance for CKD classification and identify the most effective approach[75–77]. Furthermore, ensemble stacking models were constructed by integrating[76,78] XGBoost, RF, LR, and Neural Networks, which demonstrates the highest performance. Additionally, Shapley Additive exPlanations (SHAP) were employed to analyze the contribution of each input feature to CKD classification. Unlike other traditional feature importance metrics, SHAP is based on cooperative game theory. It assigns each feature a value that represents its positive and negative contribution to the prediction. This approach ensures both consistency and local accuracy. It also clearly identifies key risk factors for individual patients[79]. This makes the decision-making process behind the model interpretable is important in the health sector to ensure reliability and to analyze and identify key predictors, such as those derived from health habits, lifestyle, and medical records, that serve as risk factors for targeted care and prevention[80,81]. The overall workflow of this study is shown in **Figure 1**.

We used cross-sectional national health surveys (BRFSS and NHIS) data instead of using data based on hospitals which only capture patients already receiving treatment. This dataset does not include undiagnosed cases with laboratory confirmation, but it can be used to model the complex characteristics of diagnosed individuals. Thus, this framework can be applied to identify untested members of the general population who have these high-risk characteristics and prioritize them for clinical screening. Then developed a customized Stacked Hybrid Architecture to handle the extreme heterogeneity and non-linear interactions of public health survey data. This stacked hybrid model combines the predictive strengths of XGBoost, Random Forest, and Logistic Regression and then uses a Multi-Layer Perceptron (MLP) as the meta-learner to maximize classification accuracy. We have evaluated advanced imputation techniques and demonstrated that generative imputation plays an important role in preserving minority-class data patterns. With this approach, we addressed the problem of the deletion of a substantial proportion of data from the dataset which reduces the sample size and increases the existing class imbalance. This study extensively used machine learning algorithms to ensure that there is no bias resulting from a single approach. This pipeline evaluated the complex combinations across 9 advanced imputation methods, 5 resampling techniques, and 11 different machine learning algorithms. Besides traditional clinical findings, our proposed framework identified Adverse Childhood Experience features such as parental separation as influential predictors of CKD. This study also provided the neurohormonal mechanisms connecting these psychosocial factors to renal pathology. We used SHAP to enhance model transparency and also took feedback from medical professionals. A multi-disciplinary research team, starting from data analysts, computer scientists, doctors, and professors worked on this study. This approach ensures that the extracted feature importances are not just statistical facts but also provide important information for medical experts to improve patient care and prevention.

The remainder of this study is arranged as follows. The experiment results and interpretability are discussed in section 2. Section 3 states the limitations, main contributions, and suggestions for the potential scope of future research work. Finally, section 4 demonstrates the methodology elaborately.

## Result

### Performance on NHIS datasets

The best performance of each model in classifying self-reported CKD applied to the NHIS 2020 dataset is presented in **Table 1**, ranked by balanced accuracy. It shows that the Stacked Ensemble model got the highest balanced accuracy and AUROC. **Table 2** shows the Summary of feature distribution alterations across imputation methods in the NHIS 2020 dataset. GAN, XGBoost and Diffusion perform well in the dataset to keep the original data properties. Regarding the sampling technique, RUS leads the models to the best performance in general, whereas not using the sampling technique leads to the worst performance. Details shown in **Table 3**.

**Table 4** shows that in the NHIS 2021 dataset, the ANN_2_Layer model with GAN imputation and RUS sampling achieved the highest Balanced Accuracy, Sensitivity and Specificity of 0.7537 (Cross-validation Balanced Accuracy of 0.6730 ± 0.01), 0.7397 and 0.7700, respectively. LR with XGBoost imputation and ROS sampling got the highest AUROC of 0.8069 (Cross-validation AUROC of 0.7776 ± 0.02). Besides that, the Stacked Ensemble model with GAN imputation and RUS sampling showed a balanced performance with AUROC, Balanced Accuracy, Sensitivity and Specificity of 0.7977 (Cross-validation AUROC of 0.7724 ± 0.02), 0.7516 (Cross-validation Balanced Accuracy of 0.7003 ± 0.03), 0.7362 and 0.7600, respectively.

Regarding the NHIS 2020 dataset, the Stacked Ensemble model with GAN imputation and RUS sampling achieved the highest AUROC and Specificity of 0.8229 (Cross-validation AUROC of 0.7724 ± 0.02) and 0.7661, respectively. On the other hand, GradientBoosting with GAN imputation and ROS sampling got the highest Balanced Accuracy of 0.7648 and ANN_2_Layer with XGBoost imputation and SMOTE sampling got the highest sensitivity of 0.7498. **Figure 2 (panel i)** shows the confusion matrices for the top 5 models on the NHIS 2020 and NHIS 2021 datasets. **Figure 2 (panel iii)** shows the boxplot of all sampling and models' performance on NHIS 2020. It shows that RUS and Stacking Ensemble lead to the best performance in classifying CKD.

**Performance on BRFSS datasets**

**Table 5** shows that in the BRFSS 2021 dataset, the Stacked Ensemble model with XGBoost imputation and RUS sampling achieved the highest Sensitivity of 0.7547 with overall balanced performance with AUROC, Balanced Accuracy and Specificity of 0.7960 (Cross-validation AUROC of 0.7913 ± 0.004) and 0.7256 (Cross-validation Balanced Accuracy of 0.7219 ± 0.004), respectively. GradientBoosting with GAN imputation and RUS sampling got the highest AUROC and Balanced Accuracy of 0.7977 (Cross-validation AUROC of 0.7924 ± 0.004), 0.7289 (Cross-validation Balanced Accuracy of 0.7221 ± 0.005). Again in the BRFSS 2019 dataset, the same GradientBoosting with GAN imputation and ROS sampling got the highest AUROC and Balanced Accuracy of 0.7984 (Cross-validation AUROC of 0.7936 ± 0.004), 0.7321 (Cross-validation Balanced Accuracy of 0.7239 ± 0.004). Besides the GradientBoosting, the Stacked Ensemble model with GAN imputation and RUS sampling got the highest sensitivity of 0.7751. **2 (panel ii)** shows the confusion matrix of the top 5 models of the BRFSS 2019 and BRFSS 2021 datasets.

**Model performance analysis across all datasets**

The results of the unadjusted general population cohorts (**Table 6**) show that the proposed Stacked Ensemble model always performed among the best models for early CKD screening. Although the observed PPVs of the leading pipelines ranged from 8.5% to 9.8%, this is expected according to Bayes' theorem when screening low-prevalence populations (~3.9% baseline CKD prevalence), where false-positive predictions naturally outnumber true positives. By contrast, all the top models had very high NPVs (98.5%–98.8%), meaning that the people who were predicted to be negative were more than 98.5% likely to be disease-free. Additionally, the proposed Stacked Ensemble had among the lowest false-negative rates across the cohorts evaluated (8.15 and 8.20 cases per 1,000 screened for NHIS 2021 and 2020, respectively, and 8.78–9.60 cases per 1,000 screened for BRFSS), minimizing the risk of missed CKD cases when screening. This trade-off is acceptable in clinical practice because false positive cases can then be subsequently assessed in the laboratory to confirm the diagnosis, and avoid irreversible renal decline. Lastly, the reliability diagrams (**Figure 3**) and brier scores (0.167–0.193) suggest good probability calibration over the assessed models, and the proposed framework shows a probability calibration similar to other popular models, especially in the low-probability region (0.0–0.4) that is relevant to clinical practice, supporting its suitability for real-world population screening.

**Figure 4, Tables 4** and **5** show that our custom Stacked Ensemble model appears in the list of 5 models for all 4 test sets. A combination of the Stacked Ensemble model with GAN imputation and RUS

sampling consistently shows balanced performance in three datasets and also with XGBoost, it shows a balanced performance. To further evaluate our proposed Ensemble Stacking (GAN+RUS) framework against other top-performing models, we conducted McNemar's statistical tests on both NHIS and BRFSS datasets. On the NHIS 2020 dataset, the proposed model achieved better performance improvements over all compared models (all $p<0.05$) and it outperformed Gradient Boosting with GAN+ROS ($p<0.001$) and Logistic Regression with GAN+RUS ($p = 0.015$). On the NHIS 2021 dataset, the model maintained its superiority over the majority of baselines. It outperforms models such as AdaBoost with GAN+ROS ($p < 0.001$) and Logistic Regression with XGBoost+ROS ($p<0.001$). On the BRFSS 2019 dataset, the proposed Ensemble Stacking (GAN+RUS) framework achieved significant superiority over all compared models (all $p<0.05$), outperforming AdaBoost with GAN+ROS ($p<0.001$) and Gradient Boosting with GAN+ROS ($p<0.001$). Similarly, on the 2021 dataset, the proposed Ensemble Stacking (XGBoost+RUS) configuration showed dominance over all models, exhibiting highly significant predictive improvements against Logistic Regression with XGBoost+ROS ($p<0.001$) and Gradient Boosting with GAN+RUS ($p<0.001$). These findings indicate that the proposed framework consistently outperformed the comparison models across the evaluated datasets.

We compared our multi-stage imputation, resampling, and stacked ensemble architecture to standard clinical baselines to assess its usefulness in providing meaningful diagnostic value. We tested unadjusted simple logistic regression models using six common clinical features (age, sex, hypertension, diabetes, BMI, general health status) in unadulterated survey samples (**Table 7**). The benchmark showed that simple models without imputation or data balancing were moderately successful at discrimination (AUROC 0.673 to 0.794), but ineffective as clinical triage classifiers. The natural class imbalance in general population health surveys was extreme, with a prevalence of CKD ranging from 3.9% to 7.3% across datasets. Simple baseline models were trained to predict the major class ("Healthy") for global statistical accuracy, resulting in a diagnostic sensitivity of 0.000 and a screening error rate of 39.51 to 121.14 missed cases per 1,000 cases diagnosed. Our proposed framework, combining psychosocial predictors with GAN-based imputation and powerful resampling methods, overcame this diagnostic failure. The stacked ensemble achieved a sensitivity of >0.750 across datasets, reducing missed clinical diagnoses to 7.64 and 8.20 per 1,000 screened individuals in NHIS 2021 and 2020, respectively, and 8.78 and 9.60 per 1,000 in BRFSS. Our proposed framework is computationally complex, but it drastically improves and balances the performance of CKD classifications.

**Feature importance analysis using SHAP values**

In **Figure 5**(b) for the BRFSS 2021 dataset, SHAP analysis using Stacked Ensemble models (GAN filling and RUS sampling) provides a comprehensive view of the factors influencing CKD classification. Among all the features it is seen that difficulty in walking, blood pressure and age with SHAP values of 0.19, 0.15 and 0.14 are the top influencer in the CKD classifications. Difficulty Walking represents severe physical frailty and systemic fatigue, which are disabling physiological consequences that frequently consider as CKD advances. The next most impactful factor is blood pressure, which makes sense because high blood pressure is a well-known risk factor for CKD. Here, 0 means no blood pressure and 1 means high blood pressure **[Table 8]**. Similarly, age is a significant factor, kidney function naturally decreases with age, putting older people at risk for kidney disease. Employment status (0.13) and physical health (0.10) also play an important role, which in turn affect kidney health **[Table 9]**. Additionally, alcohol consumption (0.05), BMI (0.02), and mental health (0.03) play an important role in kidney health, as excessive alcohol consumption, obesity, and mental health can impact kidney function. Besides that parents separation (0.02) has impact on the CKD classifications, which indicates that parental separation may increase individuals risk and has mapping between **[Table 10]**. On the other hand, some features have minimal or no impact on the model classifications. For example, verbal insult (+0.00) and health insurance do not significantly impact CKD risk.

For the BRFSS 2019 dataset, **Figure 5**(a) shows the SHAP analysis, confirming that blood pressure, difficulty in walking and age with SHAP values of 0.21, 0.17 and 0.14 are the top influencers in the CKD classifications. Beside that employment status (0.10) and physical health (0.09) again play an important role to classify kidney disease. Additionally, alcohol consumption (0.06), cholesterol awareness (0.06), BMI (0.04), mental health (0.03) and parents separation (0.03) play an important role in kidney health.

In **Figure 5**(d), SHAP analysis using the Stacked Ensemble models with XGBoost filling and RUS sampling, evaluated on the 2021 NHIS dataset, provides a comprehensive understanding of the factors influencing kidney disease classification. It further confirms that difficulty in walking, age and blood pressure with SHAP values of 0.23, 0.17 and 0.09 are the key factors in the CKD classifications. Difficulty walking often indicates chronic conditions such as diabetes, high blood pressure, or heart disease, which are considered major risk factors for CKD. Other features such as difficulty in concentration (0.16), income level (0.15), cholesterol awareness (0.12) and race (0.09) also play an important role in CKD prediction. Regular Checkup (0.08) also has a greater impact on CKD prognosis among all the features. People with regular health check-ups are more likely to get an early diagnosis of kidney disease.

For NHIS 2020, in **Figure 5**(c), SHAP analysis using the Stacked Ensemble models with GAN filling and RUS sampling, evaluated on the 2020 NHIS dataset, provides a more comprehensive understanding of the factors influencing the classification of CKD. Difficulty in walking, age, income level, difficulty in concentration and cholesterol awareness with SHAP values of 0.29, 0.16, 0.16, 0.14 and 0.13 are the top factors. Blood pressure (0.09), Regular checkup (0.08) and BMI (0.07) continue to show their impact on classification.

## Discussion

This study addresses the challenges of predicting CKD using the 2021 and 2019 BRFSS datasets and the 2021 and 2020 NHIS datasets[22,23]. CKD is a serious health problem, and many people remain unaware of their risk of developing the condition[82]. However, there are several obstacles in processing large health datasets, such as missing values, class imbalance, and high levels of data heterogeneity. This study overcomes many of the limitations of previous studies[83,84]. Nine machine learning-based techniques are used to fill in missing values, including advanced imputation methods such as GAN, Diffusion, and XGBoost. In addition, five different sampling techniques were applied to address class imbalance. At the same time,eleven machine learning models were evaluated for classification.

SHAP analysis **(Figure 5)** showed how much each of the features affected the classification of CKD risk. Previous studies[85,86] have established that high blood pressure is strongly associated with CKD risk. The possibility of CKD is very high due to irregular blood pressure. When blood pressure is high, it puts extra pressure on the kidneys. Due to excessive pressure, the kidneys cannot function properly and are further damaged[87]. Based on the interpretability of the proposed model, individuals with a history of high blood pressure appear to be highly influential in classifying the likelihood of self-reported CKD. . The risk of CKD increases significantly with age, as the functionality of various body organs decreases, including the kidney[88,89]. SHAP analysis **(Figure 5)** further confirms this finding, showing that age is among the top features in predicting CKD. These findings indicate that increasing age is associated with a higher likelihood of CKD. The analysis also found that difficulty walking (DIFFWALK) acts as a moderate to high-level risk factor for CKD. Employment status and income level were also found to be significantly associated with CKD risk, consistent with prior research[90]. Unemployment or low income causes stress, which increases the risk of hypertension and CKD. SHAP feature analysis also shows that these two have an important impact on CKD classification. In both years of the NHIS dataset, SHAP analysis **(Figure 5)** shows that regular checkup was the most important feature, which plays a key role in the classification of CKD. Regular health checkups help detect kidney problems at an early stage, and in those who do not get regular checkups, they reach a critical stage. Parental separation also contributed substantially. Although clinical markers such as high blood pressure are established risk factors for renal damage, the significant contribution of parental separation in our model's classifications aligns with existing literature proposing neurohormonal mechanisms that connect chronic stress to renal pathology. While SHAP values indicate statistical importance rather than biological causality, prior research hypothesizes that chronic mental stress and adverse childhood experiences trigger the Hypothalamic-Pituitary-Adrenal (HPA) axis, resulting in the prolonged release of catecholamines such as norepinephrine and epinephrine[91]. These hormones lead to extensive vasoconstriction and high blood pressure, both of which are recognized as key factors contributing to CKD. Moreover, imbalances in dopamine caused by stress can interfere with the reabsorption of sodium in the kidneys, and changes in serotonin levels have been associated with mitochondrial dysfunction in kidney tissues[92]. Consequently, these characteristics in our model may function not just as social indicators, but as potential proxies for physiological imbalances that impact renal health. Traumatic experiences that occur at a young age can cause chronic stress, depression, and elevated blood pressure, negatively impacting overall health and increasing the risk of CKD over time[93,94].

Moreover, in this study, a customized hybrid machine learning model has been proposed, which has been used to identify complex patterns from unbalanced data for CKD classification **(Figure 6).** While hybrid and stacked ensemble models have been used before, our designed specific architectural combination is new for this specific CKD prediction. The proposed architecture combines three complementary machine learning models with an MLP meta-learner to form a heterogeneous stacked ensemble. The design addresses challenges in the BRFSS and NHIS datasets, including missing values, class imbalance, and heterogeneous features, allowing the model to capture complex interactions and achieve higher classification accuracy for CKD. As a result, the model not only achieves higher accuracy in classification, but also efficiently understands deep relationships between complex data. Most importantly, the improvements in performance that we achieve with our customized Stacked Ensemble are not just incremental; McNemar's statistical tests showed that our model significantly outperforms the best baseline models on both of the national surveys. Specifically, the proposed ensemble showed statistically significant improvements over single Gradient Boosting ($p < 0.001$) and Logistic Regression ($p \leq 0.015$) on both the NHIS and the BRFSS cohorts. This statistical validation shows that the combination of multiple base estimators and a neural network meta-learner yields significant and non-random improvements in the diagnostic classification of models compared to the single-model approach.

In this study, we present an interpretable machine learning-based approach for CKD classification that addresses several existing challenges. Advanced machine learning techniques have been applied to fill in missing values and use sampling methods to address class imbalance issues. In addition, this study incorporates a total of eleven different models including classical machine learning, ensemble methods, ANN, Attention-based models, and a Stacked Hybrid Model to the processed dataset, which combines the decisions of multiple base models to provide robust and reliable classifications. In addition, this study includes the SHAP interpretable approach to better characterize the relationship between kidney disease risk and various features. This allows health professionals to easily identify which factors have the greatest impact on disease risk.

To evaluate the reliability of our results, we conducted an additional test in which missing values of all features were dropped directly without imputation. The results show that ACEs such as parent separation (0.03) maintained their impact even in the analysis without imputation (**Supplementary Figure 2**). This SHAP analysis shows that the impact of parent separation with imputation and without imputations is similar and there is no effect of imputation on the impact of parent separation in CKD classification. However, dropping these missing values results in a significant amount of valuable information being lost from the dataset. This limits the model's performance. Excluding data from these states risks missing some important features for making informed decisions. Our comparative analysis shows that using imputation results in relatively better model performance than dropping missing values. Therefore, we believe that imputation plays a positive role in maintaining data utility and model stability in this context. At the same time, we acknowledge that this subject requires further in-depth research and experimentation. Our results help to clarify a major challenge of handling missing data when the target disease is rare in the overall dataset. Studies show that using traditional imputation methods on imbalanced data affects the model's ability to predict the rare cases correctly[95]. Using GANs[96] allowed us to keep the original patterns of the minority CKD class far more effectively than traditional imputation techniques. Furthermore, by keeping our cross-validation strictly inside the loop, we completely avoided "data leakage." This ensured our performance scores are reliable, avoiding a common mistake that often invalidates predictive models in healthcare[70,97,98].

Overall, this study provides an interpretable, and cost-efficient solution to large-scale CKD classification. This provides interpretable models that address important issues, including the complexity of feature selection, the challenge of filling in missing values, and more. Furthermore, using SHAP analysis, this study is able to analyze the contribution of each feature, which helps health professionals understand and trust the decision-making process behind the models, and identify key risk factors for targeted patient outcomes.

Neurohormonal, demographic, lifestyle, and socioeconomic factors collectively play a complex role in the development and progression of CKD. Prior literature suggests that physiologically, the brain interacts with the renal system through the hypothalamic-pituitary-adrenal axis and neurotransmitters. For example, excess dopamine has been hypothesized to trigger hypertension[99], while altered serotonin levels lead to renal mitochondrial dysfunction[100]. These biochemical pathways are heavily affected by emotional stress such as parental separation[101], verbal abuse, and psychological distress which raise blood pressure and speed up the cycle of kidney failure[102]. While aging itself is not a direct reason for CKD, the rate jumps from 12% in adults aged 45-64 to 34% in those over 65. This is mostly because older adults normally have other overlapping health issues[103]. Furthermore, sex hormones influence disease trajectory such as protective estrogen, which leads to a higher CKD prevalence in females[84], whereas detrimental testosterone is responsible for significantly higher mortality rates in males[5,84]. Due to unequal social and healthcare barriers, Black, Hispanic, Asian, and Native American communities are more vulnerable to suffering from kidney disease[104]. Unhealthy habits, including poor diets, not drinking enough water and taking too many painkillers directly affect the kidneys[105], while smoking causes severe cardiovascular and kidney problems[106]. Furthermore, high body weight, a lack of exercise, and heavy drinking indirectly speed up kidney damage by causing other metabolic and liver problems. Lastly, social struggles such as limited education[107], housing problems, unemployment, and no health insurance also prevent people from learning about their health and getting early medical care, leading to much worse results.

A study on Central America and Mexico in 2017 found that 11.9% of the total population had CKD. Among them, women are slightly more affected (12.3%) than men[84] (11.5%). That same year, CKD was responsible for 7.6% of all deaths in the region, making it the second leading cause of mortality. From 1990 to 2017, over nearly three decades, the CKD death rate jumped by a massive 60.9% (an average of 42.1 deaths per 100,000 people). Countries such as El Salvador are seeing rates as high as 71.4 per 100,000[5,84]. The overall impact of the disease (measured by loss of healthy life) hits countries like El Salvador, Mexico, Nicaragua, and Guatemala the hardest. Men suffer substantially more regarding this than women. Moreover, the relation between liver damage and kidney disease is severe. Studies show that among patients with liver cirrhosis, 70% developed AKI, 17% had both AKI and CKD, and 13% suffered from CKD[107].

There are still some limitations. This study relies on self-reported CKD rather than laboratory-confirmed diagnosis based on clinical records, estimated glomerular filtration rate (eGFR), or albuminuria. Therefore, our model classifies the likelihood that an individual self-reports a kidney disease diagnosis, rather than detecting the true physiological onset. Because early-stage CKD is largely asymptomatic, models trained on self-reported survey data primarily capture individuals who have already interacted with the healthcare system and received a formal diagnosis, potentially limiting the model's ability to identify truly undiagnosed, early-stage cases. Careful feature selection and preprocessing were applied to reduce such bias. In terms of study design, it's important to know that BRFSS and NHIS are cross-sectional surveys that record features and results at the same time. So, the main goal of our framework is to accurately classify disease status and find important risk factors. This method works well to find strong statistical links that can be used for instant risk stratification and screening. Another limitation of our cross-sectional design is the potential for reverse causality. Some of the most highly ranked predictors (difficulty walking, physical health status, regular checkups, and difficulty concentrating) may work as proxies for existing comorbidities, general frailty, healthcare utilization, or downstream consequences of illness, rather than true upstream risk factors. We kept these features in our framework as we wanted to do concurrent case finding and opportunistic risk stratification. In the real world, if a clinician sees these broad indicators of disease or multiple health care visits, it may be a practical indicator for starting targeted screening for CKD. This, however, does not allow the model to be used as a prospective early-warning system using only upstream etiologies. Due to the high class imbalance in the dataset, Precision and F1-score were not prioritized as primary evaluation metrics[108,109]. Instead, we prioritized Balanced Accuracy and AUROC, which provide more reliable evaluations in this case. As the disease is relatively rare in the general population, Precision (PPV) is significantly impacted by the base rate fallacy and is not as effective as Sensitivity at predicting screening efficacy[110]. That's why we focused on Recall to minimize False Negatives, ensuring that individuals who require screening are not overlooked. The research was conducted using only traditional ML models (such as LR, RF, XGBoost, stacking, and advanced techniques like neural networks). However, Large Language Models (LLMs) or other advanced AI methods were not explored, which could increase classification power and provide better results. To overcome these limitations, future work will involve more datasets, LLMs, and other advanced AI models. This study used the unweighted survey data, without applying the complex sampling weights and strata provided by both the BRFSS and NHIS datasets. Thus, the performance measures, baseline characteristics, and SHAP results reflect the performance of this specific analytical cohort, not nationally representative estimates. Future research could look into using survey weights in ensemble machine learning models. Evaluating a wide range of imputation and resampling methods helps identify the most effective pipeline for this dataset. We utilized strict 5-fold cross-validation and a completely isolated test set to carefully address the chances of model-selection optimism. This ensures our findings are robust and reliable. Future research can look forward to validating these optimal configurations in external cohorts to fully establish the model's broader generalizability.

In conclusion, our framework can be useful as a practical tool for real-world healthcare scenarios. For public health planning, it will help policymakers to identify vulnerable groups such as those affected by childhood trauma. So, they can use their resources where they are needed most. For CKD screening, it provides a low-cost, population survey-based tool to identify individuals who closely match the risk profiles of diagnosed patients, serving as a triage system to prioritize high-risk individuals for confirmatory laboratory testing. We perform computational profiling on the proposed GAN-imputed Stacked Ensemble using various resampling methods. With Random Undersampling (RUS), the ensemble was able to complete full training on the BRFSS 2019 dataset (333,328 baseline records; compressed to 26,042 balanced records) in 32.71 seconds with a serialized disk footprint of 97.49 MB. With full training (ROS 640,614 records), it took 743.85 seconds with a 471.63 MB footprint. Similarly, training time and memory usage were different for each technique on the NHIS 2020 dataset: GAN+RUS used 2.31 seconds and 4.47 MB, SMOTE used 33.85 seconds and 25.49 MB, and ADASYN used 36.53 seconds and 25.41 MB. In all configurations, the Stacked Ensemble achieved an inference latency of 0.018 to 0.025 milliseconds per patient (>38,000 to >54,000 evaluations/second), making it compatible with typical Electronic Health Record (EHR) systems without the need for dedicated GPU infrastructure. Finally, for clinical decision-making, our framework gives doctors clear, personalized risk profiles for each patient. It will help them create targeted, early treatment plans rather than relying on generic screening.

## Methods

An early classification model for CKD was developed using large-scale health survey data, following a multi-stage workflow, as detailed in **Figure 1**. Feature selection, feature mapping, and advanced imputation were applied to handle missing values. Because generative methods such as GANs, Diffusion, and Autoencoders output continuous numeric values, a post-imputation constraint step was applied. All imputed continuous values for categorical and ordinal survey variables were rounded to the nearest integer to ensure they mapped back to the valid discrete categorical codes defined by the CDC codebooks. Resampling techniques addressed class imbalance, followed by model training and evaluation. SHAP was used to determine the most influential risk factors. Each stage is described in detail in the following subsections.

### Dataset description

Our study utilizes two widely used public datasets, BRFSS[111] and NHIS[82]. We specifically work on BRFSS 2021, BRFSS 2019, NHIS 2021, and NHIS 2020 dataset. The datasets are collected by telephone surveys and include information on health behaviors, conditions, and socioeconomic factors of the people of the United States[112]. The first two datasets comprise 438,693 participants with 304 features and 418,268 participants with 342 features, respectively, while the last two datasets contain 29,482 participants with 622 features and 31,568 participants with 617 features. However, the datasets are imbalanced and contain a significant amount of missing data. **[Table 11]** illustrates the information about missing values for the BRFSS 2021 dataset. [**Table 12]** presents the baseline sociodemographic and clinical characteristics of the population, stratified by kidney disease status.

### Feature selection

Exploring several literature about CKD and chronic health risk factors[113,114], we selected 32 input features from distinct categories based on health status, adverse childhood experiences, demographic and disability information which are enlisted in **[Tables 8, 9, 10 and 12]**. Among the 32 variables in the BRFSS dataset, 28 are categorical and 4 are numeric which is demonstrated in [**Table 11**]. For the NHIS dataset, we excludes 11 features and we mark the common features. The characteristics of the features are parental separation, physically injured, depressed partner, verbally abused, sexually touched, sexual harassment, basic needs, drug-addicted family member, **age, race, gender, education level, marital status, smoking status, alcohol consumption, physical activity, income level, regular checkup,** eating habits (fruits), eating habits (vegetables), **housing status, deaf**, **BMI**, **blindness, difficulty concentrating, difficulty walking, health insurance, mental health,** physical health, **employment status, blood pressure, and cholesterol awareness**.

### Data validation and splitting

To maintain strict data integrity and completely prevent data leakage, the dataset was partitioned using an 80/20 stratified train-test split prior to any missing value imputation. During this step, the target variable (Kidney Disease) was excluded to ensure missing values were inferred solely from the relationships among independent features. The imputation algorithms (e.g., GAN, Diffusion, Autoencoders, XGBoost) were strictly fitted only on the 80% training set. The remaining 20% unseen test set (83,333 samples) was kept completely isolated and was subsequently transformed using only the parameters and synthetic distributions learned exclusively from the training data. Given the high computational cost of our proposed pipeline, we performed 5-fold cross-validation strictly within the bounds of this isolated, pre-imputed training set. Crucially, methodological literature heavily cautions against applying resampling or scaling prior to cross-validation, as it leaks distributional information from the validation folds into the training process, resulting in severe "optimism bias" and artificially inflated performance metrics [97,115]. This ensures that scaling parameters and synthetic data samples were generated based solely on the training folds before being applied to the validation folds. All resampling methods (ROS, RUS, SMOTE and ADASYN) were applied only to the training folds to ensure the evaluations accurately reflect the real-world clinical screening utility and not artificially created lab conditions. The held-out test sets were strictly kept in their original, naturally imbalanced epidemiological distributions (approximately baseline CKD prevalence of ~3.9%). All population-level screening metrics (Positive Predictive Value (PPV), Negative Predictive Value (NPV), per-thousand screening error rates, and probabilistic Brier scores) are unbiased estimates of real-world deployment performance when measured against this unadjusted baseline. We also conducted McNemar’s statistical test against baseline models to establish the superiority of our best models.

### Data preparation

The primary outcome for this study was self-reported CKD. In the BRFSS dataset, this was based on the survey question: 'Not including kidney stones, bladder infection or incontinence, were you ever told you had kidney disease?'. In the NHIS dataset, the question to get this feature value was 'Have you EVER been told by a doctor or other health professional that you had ...Weak or failing kidneys?' with the additional instruction to 'Do not include kidney stones, bladder infections, or incontinence'. For both datasets, positive responses were stored as 1 (CKD) and negative responses as 0 (Healthy). Any missing, 'Don't know/Not sure,' or 'Refused' responses for this specific target variable were dropped to ensure ground-truth certainty for model training.

In the BRFSS and NHIS datasets, missing or invalid data, such as responses coded as '7', '9', '77', or '99' (indicating 'don't know/unsure' or 'refused'), as well as structural missingness caused by survey skip patterns, were uniformly converted to NaN. Rather than manually applying any rules to fill missing values, all NaN values were passed directly to the advanced imputation algorithms. It allows the models to dynamically fill the most statistically probable value based on the respondent's overall feature distribution. Furthermore, variables with a natural order, such as age group and educational level, were preprocessed, such that age 18-24 was coded as 1 and increased sequentially. Binary variables, such as “no” and "yes", were recorded as 0 and 1, respectively. This preprocessing step ensured uniform representation and comparability of all features.

The features selected from the BRFSS 2019 and 2021 datasets were organized into four main categories. **Table 10** shows the Adverse Childhood Experiences (ACEs), **Table 12** presents demographic characteristics, **Table 13** lists activity- and diet-related features, and **Table 8** includes disability- and health-related variables. To improve clarity and facilitate interpretation, each original feature (e.g., ACEDIVRC, _AGEG5YR, SMOKE100, DEAF) was assigned a descriptive new name (e.g., Parents Separation, Age, Smoking Status, Hearing Difficulty) along with a brief description explaining the survey question or the meaning of the feature. Each feature was carefully grouped and mapped into specific categories to ensure it is easy to understand and use.

### Missing value generation

We chose our imputation methods based on the reason data was missing in the first place, whether it was Missing At Random (MAR) or Missing Not At Random (MNAR)[65]. Traditional statistical methods usually assume that data follow simple, straight-line patterns. However, massive and complex medical datasets rarely follow these strict rules[54]. Because of this, we decided to use advanced imputation techniques. These advanced techniques can easily handle complex data patterns without forcing the information into strict mathematical boxes[55,96] .

**Forward-Backward** filling is a basic missing value technique in which missing data is replaced with the closest valid value[116]. Forward filling uses the previous value and backward filling uses the next value. Although simple, this approach has limitations in survey data because it only considers one variable and does not consider correlations with other variables. We are using this technique as a baseline filling method. **KNN** imputes missing values by averaging the values of the nearest neighbors based on a defined distance metric, such as Euclidean distance or cosine similarity[117]. KNN creates challenges when working with large datasets. To overcome this limitation, we divided the dataset into 5 chunks containing at most 100,000 samples[118]. But KNN is not used in our final framework because of its performance and time constraints. **Autoencoder** being a type of ANN base model, an autoencoder is a powerful technique for predicting missing values, especially when dealing with large datasets. It is also used for data compression, noise reduction, and feature learning[119]. This approach works through two functions: an encoding function that transforms input data, and a decoding function that recreates input data for output. Working with missing value fill, the autoencoder is trained by preprocessing the data through the encoder and then the decoder. Best scalable, it can handle large and complex datasets efficiently[119]. **XGBoost** sequentially builds decision trees, correcting errors in previous ones with each new tree[120]. It uses L1 and L2 regularization to reduce overfitting. To handle missing data, XGBoost detects missing values, separates incomplete rows from the training data, and trains the model on the complete data. This process enables XGBoost to handle incomplete datasets effectively while maintaining robust performance[121]. The **diffusion** model consists of a forward step, where random Gaussian noise is added to the data and a reverse diffusion step where the original data is recovered by denoising process[42]. Despite having a good reputation for generating flexible and high-quality data, the diffusion model is very slow due to the high number of steps while sampling. Our model uses a neural network with three fully connected layers, accepting input based on the number of columns and a hidden layer size of 128. It applies ReLU activation to the two hidden layers and outputs imputed values with the same magnitude as the input[122]. **GAN** consists of a generator that tries to duplicate the original data and the discriminator tries to determine whether it is original or duplicate[122]. GAN can fill in missing values based on the entire dataset. It also fills in missing values while keeping a close matching to real data. In this framework, the generator observes some components of a real vector and attempts to impute the missing components. Meanwhile, the discriminator attempts to distinguish between observed and imputed components. This adversarial approach ensures that the imputed values maintain the actual pattern of the data[96]. We defined two neural networks for generator and discriminator with ReLU activation function. Adam optimizers with a learning rate of 0.001 and binary cross-entropy loss function are used. The model is trained for 400 epochs. After training, the generator is used for missing data and replaces only the missing values[123]. **MF** is a well-known technique in collaborative filtering-based models. It computes complex matrix operations and reconstructs a matrix by reducing an original matrix into subparts[124,125]. Basically, it is used in data compression, feature learning, topic modeling, image recognition, and movie recommendation systems. In handling process data to fill in missing values, the MF model has a sensitive method SVD assigned, which works for balanced numbers of observed and missing entries. It works efficiently when processing dense matrices[126]. **Mean** imputation is a straightforward approach that can easily implement unknown values based on the existing data statistics. This approach ensures missing values are filled with the average value of the non-missing data points. The **drop** method removes rows with missing values from a dataset, and it is the most straightforward way to handle missing values.

**Resampling techniques**

As the dataset is highly imbalanced, with only 3.9% positive cases, this imbalance could induce bias in the machine learning models, affecting their ability to generalize effectively. Algorithms usually bias themselves towards the majority class as they are designed to minimize global error. In an imbalanced dataset, classifiers consider the minority class as outliers to get high overall accuracy which leads to poor sensitivity. Therefore, applying sampling strategies is important to balance the distribution so that algorithms can effectively learn the decision boundaries for the rare class[71]. To address this issue, we implemented 4 resampling algorithms. **ROS** balances the dataset by randomly duplicating the minority class, without changing the majority class[127]. **RUS** balances the dataset by reducing the size of the majority class. It randomly removes instances from the majority class, effectively decreasing its influence in the dataset[128]. Unlike simple oversampling, which thoroughly makes duplicates of existing records. This increases the risk of overfitting. SMOTE generates fully new synthetic data samples by interpolating between a minority sample and its nearest neighbors. This balances the distribution of the classes and helps the model generalize better to unseen data[72,129]. This difference is then multiplied by a random value between 0 and 1 and added to the original instance, generating a new synthetic data point[72]. **AdaSyn** builds upon SMOTE by focusing on hard-to-learn instances within the minority class[130]. It calculates the density distribution of minority class samples, identifying those that are harder for the model to classify due to their sparsity in the feature space. AdaSyn generates more synthetic samples for these challenging instances by interpolating between the selected sample and its nearest neighbors, resulting in a more intelligently balanced dataset[74]. Overall, these techniques collectively aim to mitigate the challenges of imbalanced datasets by improving representation, enhancing generalization, and reducing bias in the resulting ML models.

Different resampling and missing value handling methods were applied to the training set, and the results are presented in **[Supplementary Tables 7, 8]**. Without resampling, Class 0 made up about 96% of the data, while Class 1 accounted for only 3–4%. When techniques such as ROS, RUS, or SMOTE were applied, the class distribution became nearly balanced at around 50:50 for most missing value methods. AdaSyn slightly improved the representation of the minority class, making it better represented in the dataset.

## Predictive Modeling Architectures

This study implemented eleven predictive modeling approaches to evaluate and classify CKD systematically. These architectures were selected to handle the high dimensionality, class imbalance and complex pattern of data. To ensure a comprehensive analysis, this study explored classical machine learning, deep learning and a custom stacked hybrid architecture. We trained models on the train set of BRFSS 2019 and evaluated on both BRFSS 2019 and BRFSS 2021. Also trained models on the train set of NHIS 2020 and evaluated on both NHIS 2020 and NHIS 2021. Architectural details and hyperparameter settings for all predictive models are shown in **Table 14**.

### Classical machine learning models

**LR** is typically used in medical and health domains because of its ability to model the relationship between dichotomous dependent variables and one or more independent variables. LR analyzes the relationship between multiple independent variables and a categorical dependent variable and then estimates the probability of occurrence by fitting the data to a logistic curve[131]. LR fails to predict correctly when it has many additive coefficients[132]. **DT** is typically used in healthcare, finance, e-commerce because it provides a process by taking simple results by tree structure[133]. It starts from the root node and splits the data. This process repeats at each node until the end is reached, where the leaf node provides a final decision. Sometimes overfitting can occur with small datasets. **XGBoost** is mainly used when working with high-dimensional data and the need for high accuracy. It builds trees sequentially, with each new tree trying to correct the mistakes of the previous tree[120]. XGBoost automatically handles missing values and finds out the best way to handle missing data[134]. However, it suffers high complexity, overfitting, and high memory usage. **AdaBoost** is typically used to improve the accuracy of a given model by merging many "weak" trainees[135]. Initially, AdaBoost starts training a weak classifier and then weights it based on how well it performs[136]. However, this algorithm may be sensitive to noisy data and a slow training process. **RF,** which is typically used for classification and regression, utilizes multiple decision trees on different subsets of data[137]. Each tree contributes a "vote,” and the majority of the average decision is selected for classification and regression, respectively, which reduces the risk of overfitting and enhances the overall performance. However, RF has high computational cost and memory usage, which may not perform well on sparse datasets. **Gradient Boosting** can be used in medical datasets because these datasets often have class imbalances and it can handle complex patterns[138]. It works well with structured data, which makes it appropriate for patient risk assessment or diagnostic forecasting. The gradient boosting model works by constructing different machine learning models sequentially. It enhances overall performance by iteratively adding models to fix the mistakes caused by earlier models[139]. Though there are some limitations, like training can be slow when we work with large datasets like BRFSS. Without handling missing values, it can't give the right accuracy[140]. **LightGBM** is used in the finance, health domains and developed predictive models of structured data[141]. It builds a decision tree that develops leaf wise which uses a histogram-based method for efficient data processing. Sometimes overfitting can occur with small datasets.

**Deep learning architectures**

Artificial Neural Network (ANN) is inspired by the human brain and how it processes information. ANN gathers knowledge by identifying patterns and relationships in data and learns through experience, not from programming[142]. We designed three ANN models with different architectures. A single-layer ANN containing 64 neurons and a two-layer ANN containing 128 and 64 neurons. We used ReLU as an activation function in the hidden layers and sigmoid activation function in the output layer. We also created another version of ANN model of two layers [64, 32] with an attention layer which essentially contains significantly more parameters compared to the previous ANN models. To help the model better understand the data and easily identify patterns, we added Attention layer directly to the inputs before they pass through the Dense layers. By doing this, the model can automatically prioritize the most important variables from the noisy data while ignoring irrelevant details. This focused, weighted information is then passed through the remaining hidden layers to make the ultimate CKD classification[143].

**Hybrid model implementation**

We designed the hybrid machine learning model as shown in **Figure 5**, two stages to combine the strengths of different models. The main objective is to reduce the generalization error by using a meta-learner in level 1 to intelligently combine the predictions of multiple heterogeneous base models in level 0 models. By learning the patterns of making errors of base models, the meta-learner corrects and improves its biases to produce a final classification that is more accurate than any single model[144]. Using multiple models helps capture different patterns in the data, and the neural network meta-learner combines their predictions to improve overall accuracy and reliability.

**Base Estimators:** The first stage contains 3 different machine learning models which are XGBoost, Random Forest, and Logistic Regression. These 3 models were trained independently on the training dataset so that each model can learn and explore the complex patterns of the data based on its strengths.

**Meta-Learner:** In the second stage, an MLP is utilized as the meta-learner. The meta-learner was trained using out-of-fold predictions from the base estimators during 5-fold cross-validation on the training set to prevent overfitting. Hyperparameters for both the base estimators and the meta-learner (detailed in Table 14) were selected based on preliminary empirical validation strictly within the training folds. The final 20% test set was kept completely untouched throughout hyperparameter tuning and model training. It was utilized only once at the very end to evaluate the fully finalized stacked ensemble.

Existing literature on the classification of CKD often adopts the conventional homogeneous ensemble methods like bagging or boosting (Random Forest or XGBoost), but the use of heterogeneous stacked ensembles is limited[36,38,42]. Moreover, the use of stacking in wider medical contexts usually involves simple linear models (such as Logistic Regression) as meta-learners. Our proposed architecture is quite different, using a Multi-Layer Perceptron (MLP) as the meta-learner. We use a variety of base estimators that learn linear and non-linear patterns (Logistic Regression, Random Forest, XGBoost) to get a full predictive feature space. The MLP meta-learner then uses non-linear transformations to these out-of-fold predictions. This enables the ensemble to learn very complex decision boundaries and correct very complex error patterns that standard ensembles or linear meta-learners fail to learn, making it very well suited to the non-linear interactions found in large-scale public health survey data.

**Model interpretability using SHAP**

We used SHAP to interpret our ensemble's predictions and to analyze feature contributions. We chose SHAP as our standalone explainability framework over heuristic tree-based metrics (such as Gini/Gain) and approximation methods (such as LIME), which may have inconsistent and unstable performance around complex decision boundaries[145]. Based on cooperative game theory, SHAP ensures local accuracy and consistency. Importantly, it consistently distributes credit over correlated socioeconomic and clinical factors that are built into epidemiological data, enabling strong individual and population-level insights.

## Data availability

We used the publicly available Behavioral Risk Factor Surveillance System[22,23] (BRFSS 2021 and BRFSS 2019) and the National Health Interview Survey[82] (NHIS 2021 and NHIS 2020). As these datasets are publicly available without any requirements, no IRB is required for this study.

## Code availability

All analysis conducted for this study was performed using Python along with the latest versions of libraries available at the time of submission (2025). All the experiments were conducted on Google Colab platform. The code is publicly available in Github: https://github.com/AtikShams/CKD.

## Acknowledgements

The authors declare that no funds, grants, or other support were received during the preparation of this manuscript.

## Author contributions

T.P. conceived and designed the study. S.F., M.S., A.S., and T.P. prepared the dataset for experimentation. T.P. and D.M. selected the features. M.S., S.F., A.S., D.I., J.M., and T.P. conducted the experiments and analyzed the results. M.S. conducted the interpretability study, and S.M. analyzed the results from the medical perspective. A.D. worked on statistical analysis. M.S., S.F., A.S., D.I., J.M., S.M., D.M., N.A., and T.P. wrote the paper. S.A., D.E., J.F., H.I., T.H., and T.P. reviewed the study and provided strategic guidance. M.S. reviewed and revised the paper. T.P. supervised the project and coordinated the resources. All authors proofread the manuscript and provided feedback.

## Competing interests

The authors declare no competing financial or non-financial interests.

## Additional Information

The supplementary document is attached with the submission.

**Correspondence** and requests for materials should be addressed to Tanmoy Sarkar Pias.

## Figures and tables

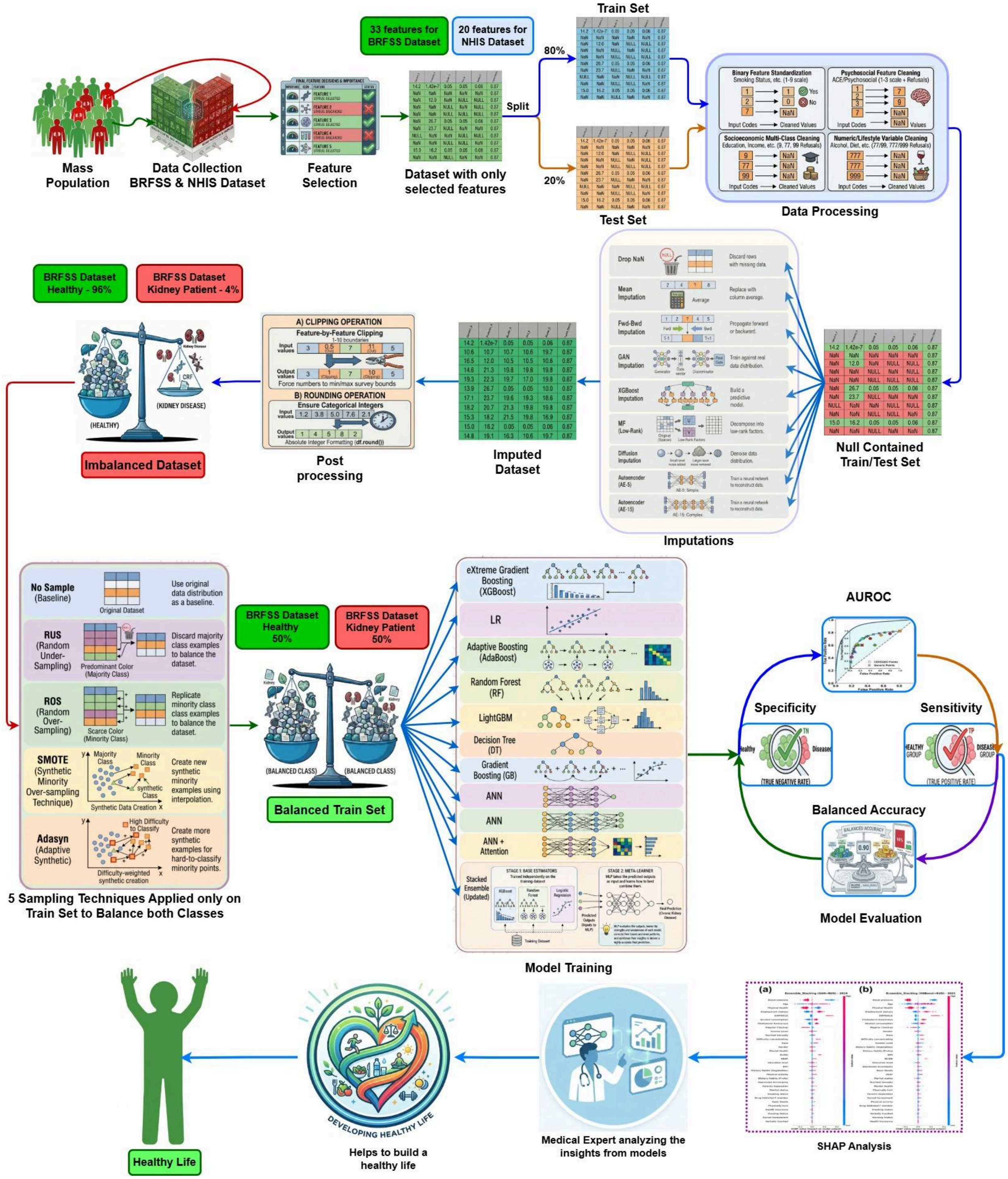

**Figure 1:** Complete workflow for chronic kidney disease classification. The figure shows the full workflow from the data collection to help patients to develop a healthy life.

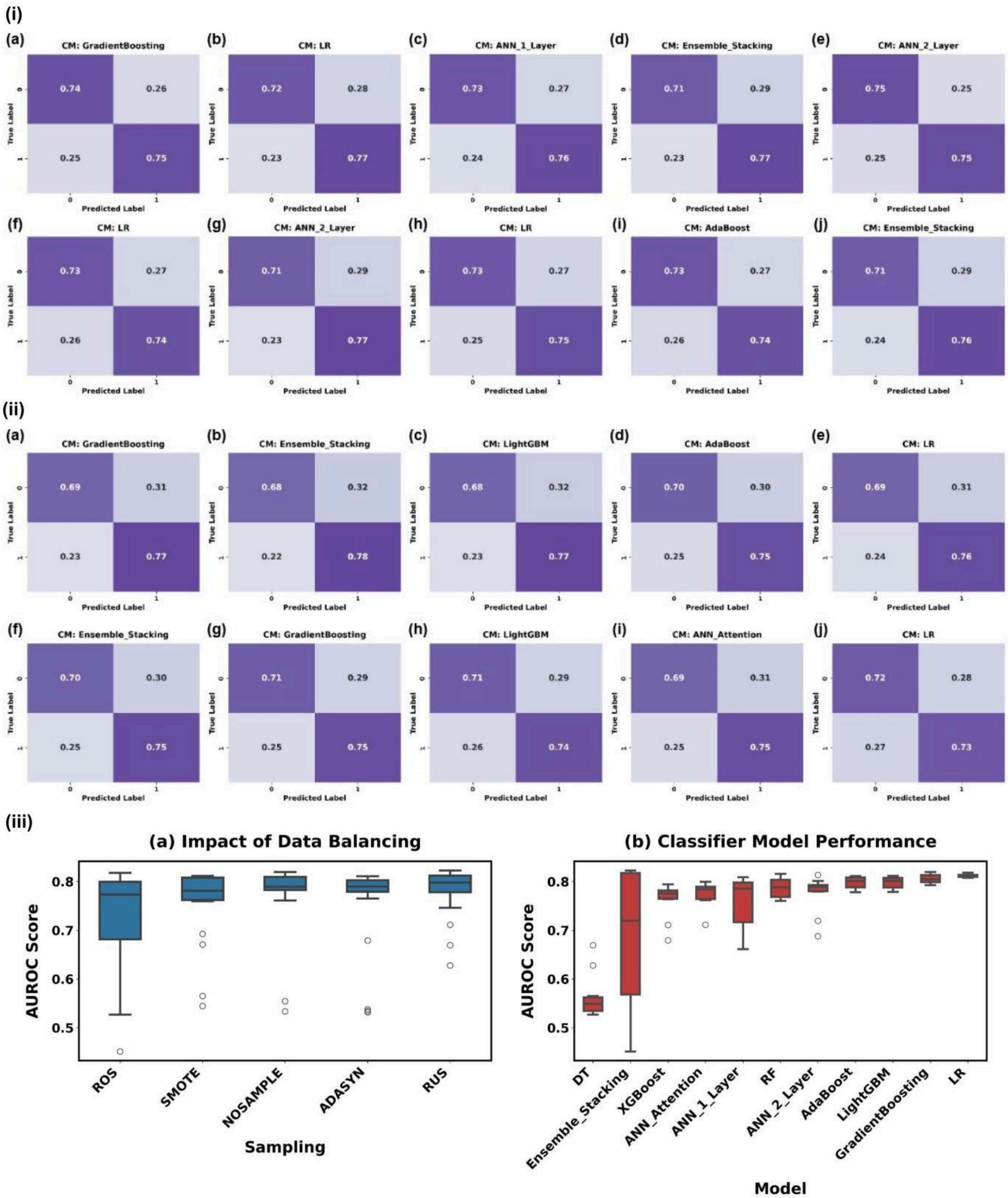

**Figure 2:** In **Panel i**, the confusion matrix shows the classification performance of the five top modeling pipelines for NHIS 2020 in the top row (a–e) and NHIS 2021 in the bottom row (f–j). Each subfigures corresponds to a specific combination of classifier, imputation technique, and data sampling strategy: (a) Gradient Boosting (GAN, ROS), (b) Logistic Regression (GAN, RUS), (c) 1-Layer Artificial Neural Network (XGBoost, SMOTE), (d) Ensemble Stacking (GAN, RUS), (e) 2-Layer Artificial Neural Network (XGBoost, SMOTE), (f) Logistic Regression (XGBoost, ROS), (g) 2-Layer Artificial Neural

Network (GAN, RUS), (h) Logistic Regression (GAN, ROS), (i) AdaBoost (GAN, ROS), and (j) Ensemble Stacking (GAN, RUS). The x-axis represents the predicted labels and the y-axis represents the true labels (0 = Healthy, 1 = CKD). Values within the matrices are normalized, where the bottom-right quadrant indicates the True Positive Rate (Sensitivity) and the top-left quadrant indicates the True Negative Rate (Specificity). Across both datasets, these optimized pipelines demonstrate strong and highly balanced diagnostic capabilities, maintaining Sensitivities between 0.73 and 0.75 and Specificities between 0.74 and 0.77. In **panel ii**, The confusion matrix shows the classification performance of the five top modeling pipelines for BRFSS 2019 in the top row (a–e) and BRFSS 2021 in the bottom row (f–j). Each subfigures corresponds to a specific combination of classifier, imputation technique, and data sampling strategy: (a) Gradient Boosting (GAN, ROS), (b) Ensemble Stacking (GAN, RUS), (c) LightGBM (GAN, RUS), (d) AdaBoost (GAN, ROS), (e) Logistic Regression (XGBoost, ROS), (f) Ensemble Stacking (XGBoost, RUS), (g) Gradient Boosting (GAN, RUS), (h) LightGBM (GAN, RUS), (i) Attention-based Artificial Neural Network (XGBoost, ROS), and (j) Logistic Regression (XGBoost, ROS). The x-axis represents the predicted labels and the y-axis represents the true labels (0 = Healthy, 1 = CKD). Values within the matrices are normalized, where the bottom-right quadrant indicates the True Positive Rate (Sensitivity) and the top-left quadrant indicates the True Negative Rate (Specificity). Across both datasets, these pipelines demonstrate strong and highly balanced performance, maintaining Sensitivities between 0.72 and 0.77 and Specificities between 0.68 and 0.71. In **panel iii**, the box plots illustrate the variance and stability of the AUROC scores across various machine learning pipeline combinations. Subfigure (a) shows that Random Under-Sampling (RUS) provides the most accurate and consistent results, whereas Random Over-Sampling (ROS) is highly unpredictable. Subfigure (b) compares the different models. It shows that algorithms like Logistic Regression, Gradient Boosting, and LightGBM perform reliably well across the board. The Ensemble Stacking model also reaches high accuracy, but it has a wider range of scores (a taller box), meaning its success depends heavily on how well the data is prepared earlier.

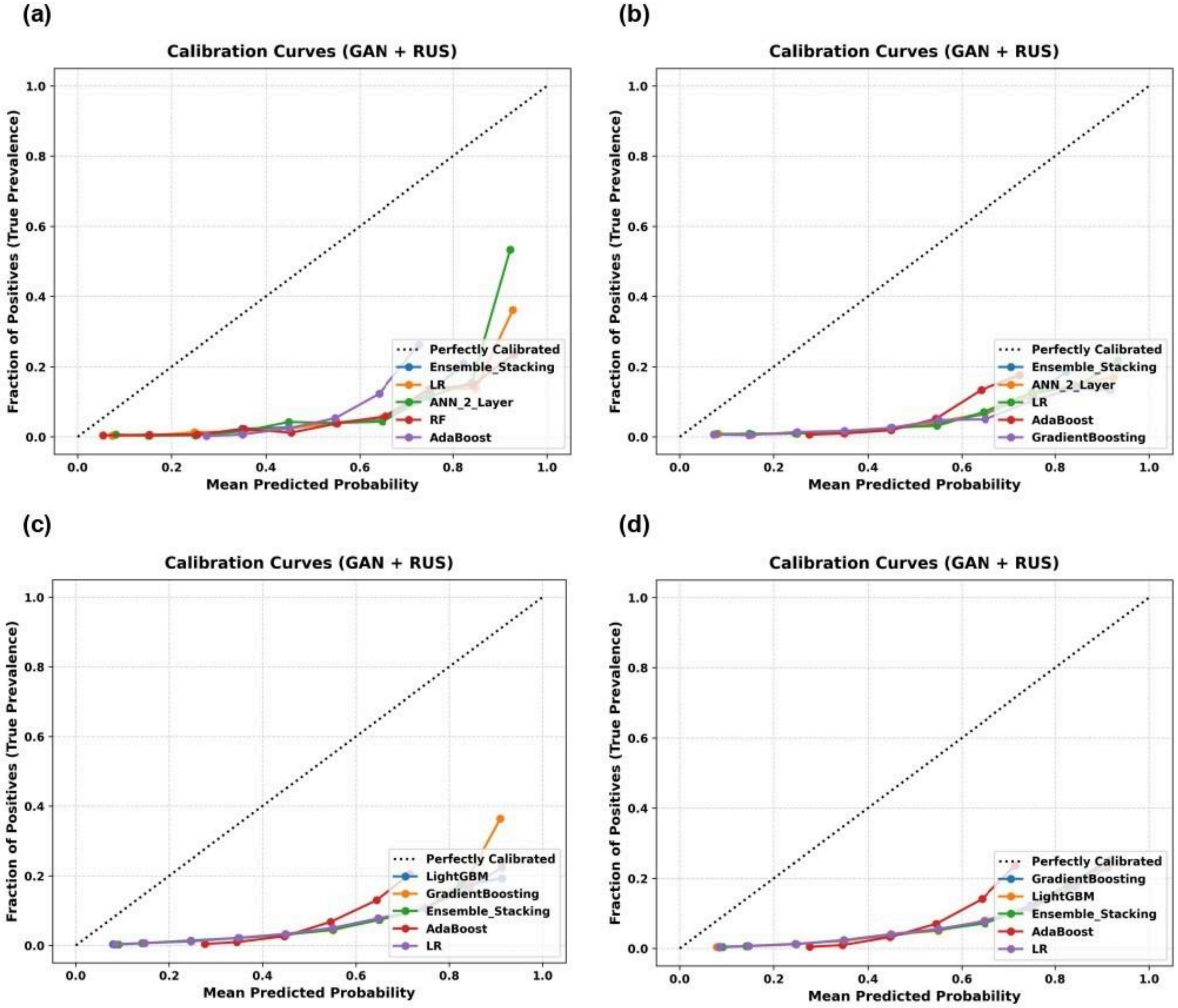

**Figure 3:** Calibration reliability of top-performing machine learning models across four datasets. Panels show calibration curves for (a) NHIS 2020 (GAN + RUS), (b) NHIS 2021 (GAN + RUS), (c) BRFSS 2019 (GAN + RUS), and (d) BRFSS 2021 (GAN + RUS). The dotted diagonal reference line indicates perfect probabilistic calibration. Curves closer to the reference line demonstrate higher trustworthiness in predicting true individual risk of chronic kidney disease.

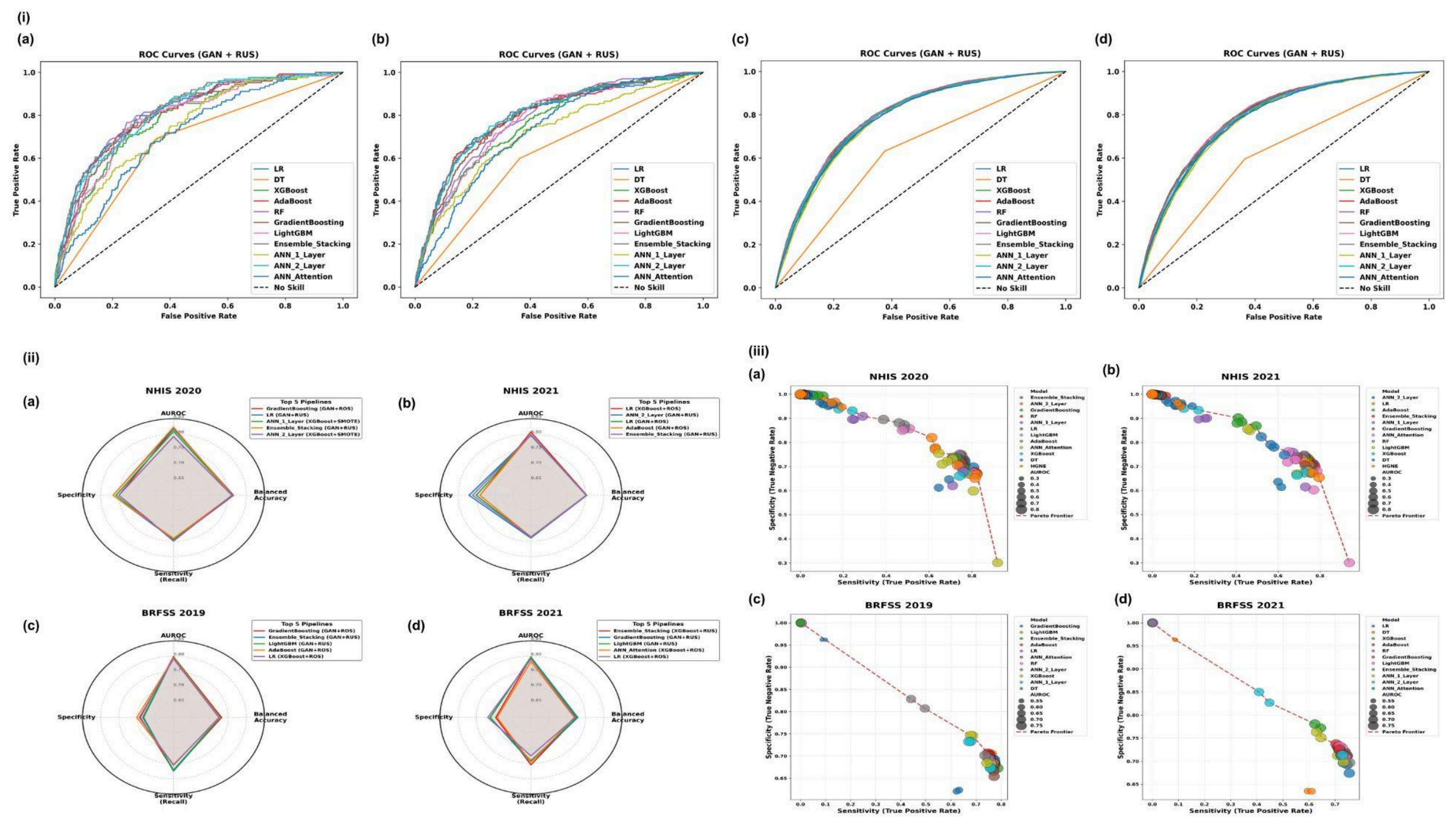

**Figure 4:** In **panel i,** AUROC curves for the tested machine learning models using the GAN + RUS pipeline. The graphs compare the predictive accuracy of different classifiers across four datasets: (a) NHIS 2020 and (b) NHIS 2021, (c) BRFSS 2019 and (d) BRFSS 2021 The curves show the trade-off between the True Positive Rate and False Positive Rate. Curves that reach higher toward the top-left corner indicate better performance, with all models performing well above the dashed "No Skill" baseline. In **panel ii,** the radar charts show the performance of the best models across four key metrics: AUROC, Balanced Accuracy, Sensitivity, and Specificity. The charts range from 0.65 to 0.85 to highlight the small differences between them. Across all four datasets (a) NHIS 2020, (b) NHIS 2021, (c) BRFSS 2019, and (d) BRFSS 2021, the models show consistently high overall accuracy (AUROC), but slightly different performance when detecting true cases and ruling out false cases. In **panel iii,** The scatter plot shows the comparison between successfully detecting kidney disease (Sensitivity) and correctly identifying healthy patients (Specificity). The red dashed line connects the absolutely optimal models. Notably, the Ensemble Stacking model shows impressive performance across all four datasets. It consistently lands right on this top-tier line with a high overall score (represented by large bubble sizes).

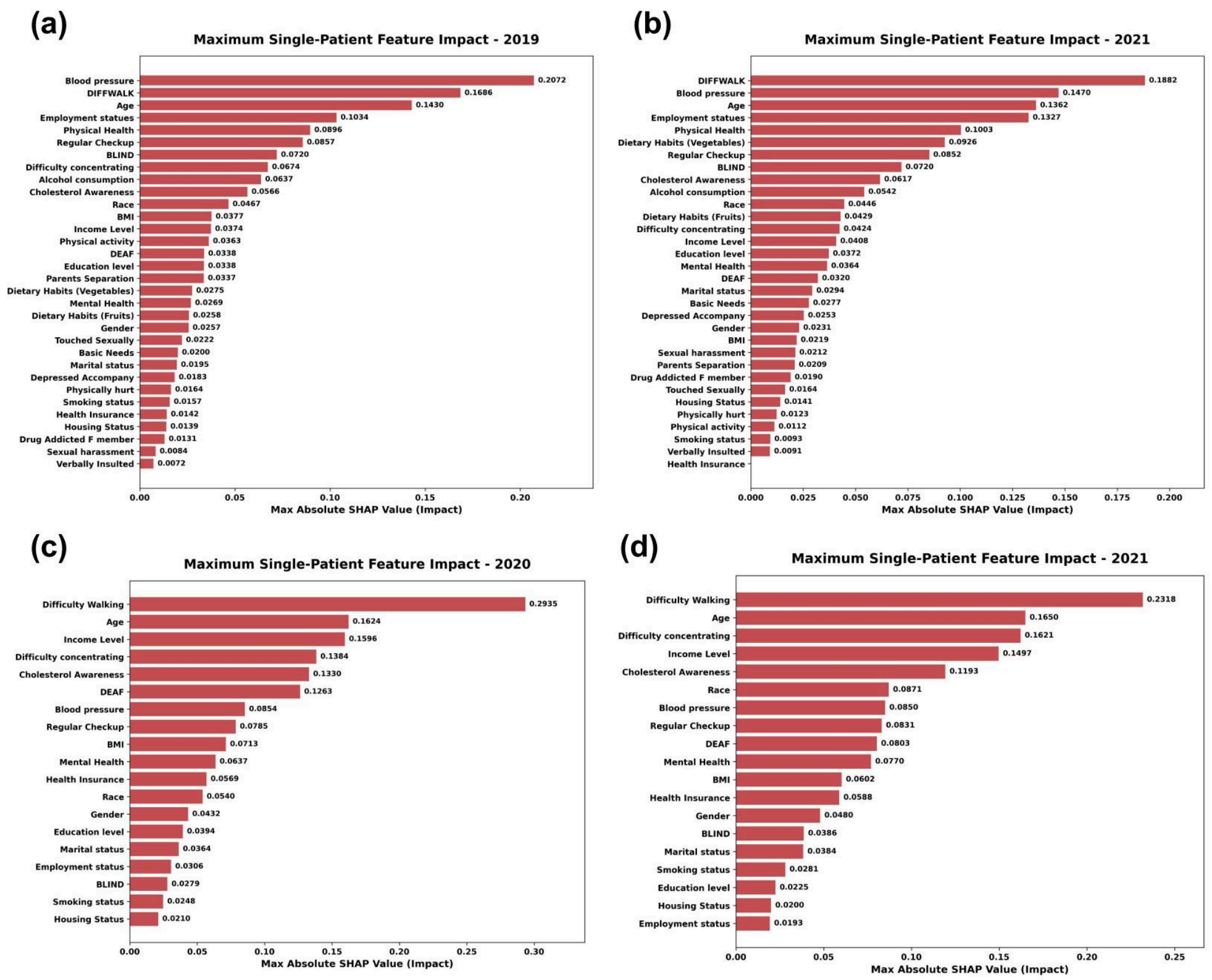

**Figure 5:** The SHAP plot shows the rank of patient features by their contributions to the model's classifications. Panels (a) and (b) show results for the BRFSS 2019 and 2021 datasets, where features such as blood pressure, difficulty walking, and age have the largest impact. Panels (c) and (d) show results for the NHIS 2020 and 2021 datasets, where difficulty walking, age, and income level are the most influential features. Longer bars indicate that a feature plays a stronger role in determining the model's final decision for an individual patient.

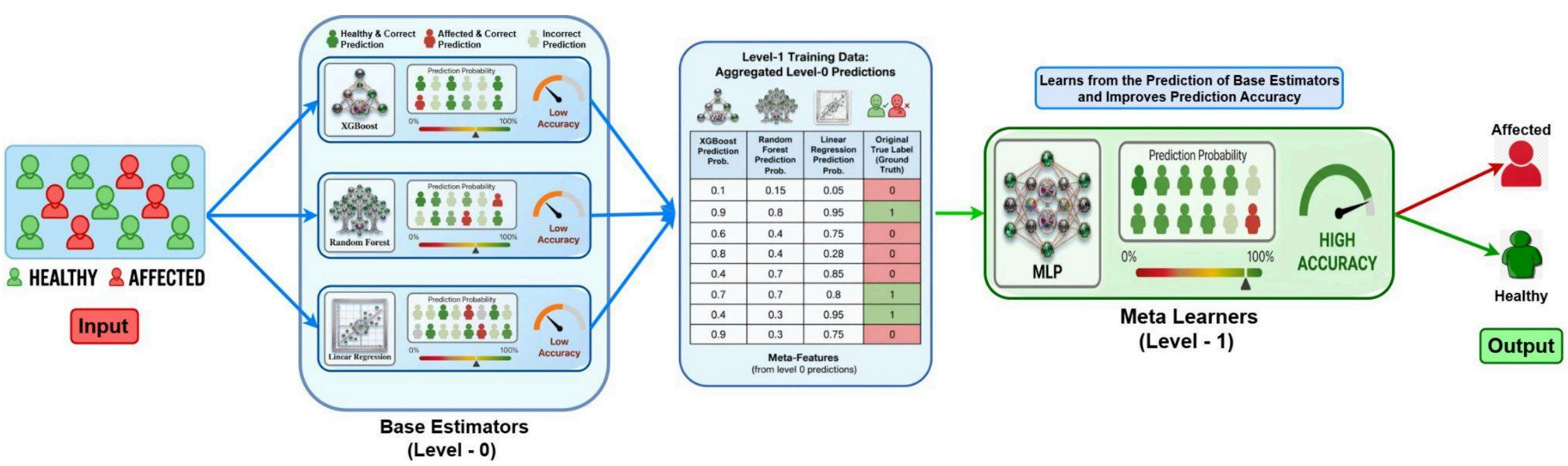

**Figure 6:** Architecture of the proposed Stacked Ensemble model combining multiple ML models for CKD classification. XGBoost, Random Forest and Logistic Regression are functioning as base estimators and MLP is operating as the meta-learners.

**Table 1:** Highest performance of each model on NHIS 2020 dataset, sorted by balanced accuracy.

| Model | AUROC | Balanced Accuracy | Sensitivity | Specificity |
|---|---|---|---|---|
| Ensemble_Stacking | **0.8222** | **0.7518** | 0.8065 | 0.6971 |
| ANN_2_Layer | 0.7918 | 0.7498 | 0.7500 | **0.7496** |
| GradientBoosting | 0.8179 | 0.7471 | 0.7500 | 0.7443 |
| RF | 0.8160 | 0.7465 | **0.8226** | 0.6705 |
| ANN_1_Layer | 0.8089 | 0.7462 | 0.7581 | 0.7343 |
| LR | 0.8184 | 0.7450 | 0.7661 | 0.7238 |
| LightGBM | 0.7790 | 0.7404 | 0.8065 | 0.6743 |
| AdaBoost | 0.8053 | 0.7364 | 0.7419 | 0.7308 |
| ANN_Attention | 0.7904 | 0.7236 | 0.7097 | 0.7376 |
| XGBoost | 0.7782 | 0.7211 | 0.7661 | 0.6760 |
| DT | 0.6696 | 0.6699 | 0.6935 | 0.6462 |

**Table 2:** Summary of feature distribution alterations across imputation methods (NHIS 2020)

| Imputation | Features Evaluated | Significant Alterations ($p < 0.05$) | Preserved Features | Rank |
|---|---|---|---|---|
| Forward-Backward* | 19 | 0 | 19 | - |
| GAN | 19 | 4 | 15 | 1 (Tie) |
| XGBoost | 19 | 4 | 15 | 1 (Tie) |
| Diffusion | 19 | 4 | 15 | 1 (Tie) |
| Autoencoder-5 | 19 | 5 | 14 | 2 (Tie) |
| Autoencoder-15 | 19 | 5 | 14 | 2 (Tie) |
| Matrix Factorization | 19 | 5 | 14 | 2 (Tie) |
| Mean Imputation | 19 | 5 | 14 | 2 (Tie) |
| Drop NaN | 19 | 7 | 12 | 3 |

**Table 3:** Highest performance of each sampling method on NHIS 2020 dataset, sorted by balanced accuracy.

| Sampling | AUROC | Balanced Accuracy | Sensitivity | Specificity |
|---|---|---|---|---|
| RUS | **0.8222** | **0.7518** | **0.8065** | 0.6971 |
| SMOTE | 0.7918 | 0.7498 | 0.75 | 0.7496 |
| ROS | 0.8179 | 0.7471 | 0.75 | 0.7443 |
| ADASYN | 0.8033 | 0.7459 | 0.7419 | 0.7499 |
| NOSAMPLE | 0.5543 | 0.5549 | 0.1532 | **0.9566** |

**Table 4:** Performance Comparison of Top 5 models using filling and sampling methods for NHIS 2021 and 2020 Datasets.

| Dataset NHIS | Model | Imputation | Sampling | AUROC | Balanced Accuracy | Sensitivity | Specificity |
|---|---|---|---|---|---|---|---|
| | LR | XGBoost | ROS | **0.8069** | **0.7537** | 0.7360 | 0.7400 |
| | ANN_2_Layer | GAN | RUS | 0.7957 | **0.7537** | **0.7397** | **0.7700** |
| 2021 | LR | GAN | ROS | 0.7986 | 0.7530 | 0.7377 | 0.7500 |
| | AdaBoost | GAN | ROS | 0.7981 | 0.7523 | 0.7370 | 0.7400 |
| | Ensemble_Stacking | GAN | RUS | 0.7977 | 0.7516 | 0.7362 | 0.7600 |
| | GradientBoosting | GAN | ROS | 0.8179 | **0.7648** | 0.7471 | 0.7500 |
| | LR | GAN | RUS | 0.8184 | 0.7633 | 0.7450 | **0.7661** |
| 2020 | ANN_1_Layer | XGBoost | SMOTE | 0.8089 | 0.7619 | 0.7462 | 0.7581 |
| | Ensemble_Stacking | GAN | RUS | **0.8229** | 0.7612 | 0.7406 | **0.7661** |
| | ANN_2_Layer | XGBoost | SMOTE | 0.7918 | 0.7603 | **0.7498** | 0.7500 |

**Table 5:** Performance comparison of the top 5 models using filling and sampling methods for BRFSS 2021 and 2019.

| Dataset BRFSS | Model | Imputation | Sampling | AUROC | Balanced Accuracy | Sensitivity | Specificity |
|---|---|---|---|---|---|---|---|
| | Ensemble_Stacking | XGBoost | RUS | 0.7960 | 0.7256 | **0.7547** | 0.6966 |
| | GradientBoosting | GAN | RUS | **0.7977** | **0.7289** | 0.7462 | 0.7117 |
| 2021 | LightGBM | GAN | RUS | 0.7964 | 0.7266 | 0.7409 | 0.7123 |
| | ANN_Attention | XGBoost | ROS | 0.7826 | 0.7194 | 0.7462 | 0.6926 |
| | LR | XGBoost | ROS | 0.7912 | 0.7220 | 0.7254 | **0.7185** |
| | GradientBoosting | GAN | ROS | **0.7984** | **0.7321** | 0.7727 | 0.6915 |
| | Ensemble_Stacking | GAN | RUS | 0.7959 | 0.7289 | **0.7751** | 0.6826 |
| 2019 | LightGBM | GAN | RUS | 0.7962 | 0.7282 | 0.7717 | 0.6847 |
| | AdaBoost | GAN | ROS | 0.7937 | 0.7272 | 0.7536 | **0.7009** |
| | LR | XGBoost | ROS | 0.7915 | 0.7249 | 0.7558 | 0.6941 |

**Table 6.** Comprehensive clinical screening utility, per-thousand error rates, and probabilistic calibration of top-performing machine learning models across national health survey datasets.

| Dataset | Model | Imputation | Sampling | PPV | NPV | FP / 1,000 | FN / 1,000 | Brier Score |
|---|---|---|---|---|---|---|---|---|
| NHIS 2021 | LR | XGBoost | ROS | 0.0885 | 0.9877 | 258.91 | 8.83 | **0.1827** |
| | ANN_2_Layer | GAN | RUS | 0.0852 | **0.9887** | 280.65 | **7.81** | 0.1936 |
| | LR | GAN | ROS | 0.0876 | 0.988 | 265.2 | 8.49 | 0.1837 |
| | AdaBoost | GAN | ROS | **0.089** | 0.9877 | **257.05** | 8.83 | 0.1963 |
| | Ensemble_Stacking | GAN | RUS | 0.085 | 0.9883 | 277.76 | 8.15 | 0.1872 |
| NHIS 2020 | GradientBoosting | GAN | ROS | 0.0963 | 0.9879 | 246.75 | 8.76 | 0.1673 |
| | LR | GAN | RUS | 0.0915 | **0.9884** | 266.53 | **8.2** | 0.1864 |
| | ANN_1_Layer | XGBoost | SMOTE | 0.0939 | 0.9882 | 256.36 | 8.48 | 0.1701 |
| | Ensemble_Stacking | GAN | RUS | 0.089 | 0.9883 | 275.01 | **8.2** | 0.1885 |
| | ANN_2_Layer | XGBoost | SMOTE | **0.0981** | 0.988 | **241.66** | 8.76 | **0.1647** |
| BRFSS 2021 | Ensemble_Stacking | XGBoost | RUS | 0.092 | **0.9859** | 291.54 | **9.6** | 0.1869 |
| | GradientBoosting | GAN | RUS | **0.0954** | 0.9857 | 277.02 | 9.93 | **0.1785** |
| | LightGBM | GAN | RUS | 0.0949 | 0.9854 | 276.46 | 10.14 | 0.1801 |
| | ANN_Attention | XGBoost | ROS | 0.09 | 0.9853 | 295.39 | 9.93 | 0.179 |
| | LR | XGBoost | ROS | 0.095 | 0.9847 | **270.51** | 10.75 | 0.1811 |
| BRFSS 2019 | GradientBoosting | GAN | ROS | 0.0924 | **0.9868** | 296.46 | 8.88 | **0.1892** |
| | Ensemble_Stacking | GAN | RUS | 0.0903 | **0.9868** | 304.97 | **8.78** | 0.1935 |
| | LightGBM | GAN | RUS | 0.0905 | 0.9866 | 303.01 | 8.92 | 0.1927 |
| | AdaBoost | GAN | ROS | **0.0929** | 0.9859 | **287.46** | 9.62 | 0.2006 |
| | LR | XGBoost | ROS | 0.0913 | 0.9859 | 293.92 | 9.54 | 0.1921 |

**Table 7.** Diagnostic performance and screening error rates of simple baseline Logistic Regression models across unadjusted national survey cohorts

| Dataset | Model | Imputation | Sampling | AUROC | Balanced Accuracy | Sensitivity | Specificity | PPV | NPV | FP / 1,000 | FN / 1,000 | Brier Score |
|---|---|---|---|---|---|---|---|---|---|---|---|---|
| NHIS 2021 | LR | None | None | 0.7396 | 0.5 | 0 | 1 | 0 | 0.9269 | 0 | 73.12 | 0.0645 |
| NHIS 2020 | LR | None | None | 0.7548 | 0.4996 | 0 | 0.9992 | 0 | 0.9286 | 0.78 | 71.32 | 0.0619 |
| BRFSS 2021 | LR | None | None | 0.6731 | 0.5 | 0 | 1 | 0 | 0.8789 | 0 | 121.14 | 0.1019 |
| BRFSS | LR | None | None | 0.7938 | 0.5 | 0 | 1 | 0 | 0.9605 | 0 | 39.51 | 0.0357 |

2019

**Table 8:** Disability- and health-related features from BRFSS 2019 and 2021.

| Feature Name Original | Feature Name (New) | Description | Value Type | Feature Map |
|---|---|---|---|---|
| DEAF | DEAF | Are you deaf or do you have serious difficulty hearing? | 1: Yes, 2: No | 0: No, 1: Yes |
| _BMI5CAT | BMI | Computed body mass index categories | 1: Underweight, 2: Normal Weight 3: Overweight, 4: Obese | 1: Underweight, 2: Normal Weight, 3: Overweight, 4: Obese |
| BLIND | BLIND | Blind or Difficulty seeing | 1: Yes, 2: No | 0: No, 1: Yes |
| DECIDE | Difficulty concentrating | Difficulty Concentrating or Remembering | 1: Yes, 2: No | 0: No, 1: Yes |
| DIFFWALK | DIFFWALK | Difficulty Walking or Climbing Stairs | 1: Yes, 2: No | 0: No, 1: Yes |
| _HLTHPLN | Health Insurance | Have any health insurance? | 1: Have some form of insurance, 2: Do not have some form of health insurance, | 0: Do not have some form of health insurance, 1: Have some form of insurance, |
| _MENT14D | Mental Health | Computed Mental Health Status | 1: Zero days when mental health not good, 2: 1-13 days when mental health not good, 3: 14+ days when mental health not good, | 0: Zero days when mental health not good, 1: 1-13 days when mental health not good, 2: 14+ days when not good |
| _PHYS14D | Physical Health | Computed Physical Health Status | 1: Zero days when mental health not good, 2: 1-13 days when mental health not good, 3: 14+ days when mental health not good, | 0: Zero days when mental health not good, 1: 1-13 days when mental health not good, 2: 14+ days when not good |
| _RFHYPE6 | Blood pressure | Have High Blood Pressure? | 1: No, 2: Yes, | 0: No, 1: Yes, |
| _CHOLCH3 | Cholesterol Awareness | Have Cholesterol Checked? | 1: Had cholesterol checked in the past 5 years, 2: Did not have cholesterol checked in past 5 years, 3: Have never had cholesterol checked, | 0: Never checked, 1: Had cholesterol checked in past 5 years, 2: Did not have cholesterol checked in past 5 years, |
| CH CKDNY2 | Kidney Disease | Have kidney disease? | 1: Yes, 2: No | 0: No, 1: Yes |

**Table 9:** Demographic Characteristics from BRFSS 2019 and 2021.

| Feature Name Original | Feature Name (New) | Description | Value Type | Feature Map |
|---|---|---|---|---|
| _AGEG5YR | Age | Reported age in five-year age categories calculated variable | 1: Age 18 to 24,2: Age 25 to 29, 3: Age 30 to 34, 4: Age 35 to 39, 5: Age 40 to 44, 6: Age 45 to 49, 7: Age 50 to 54, 8: Age 55 to 59, 9: Age 60 to 64, 10: Age 65 to 69, 11: Age 70 to 74, 12: Age 75 to 79, 13: Age 80 or older, 14: N/A | 1: Age 18 to 24,2: Age 25 to 29, 3: Age 30 to 34, 4: Age 35 to 39, 5: Age 40 to 44, 6: Age 45 to 49, 7: Age 50 to 54, 8: Age 55 to 59, 9: Age 60 to 64, 10: Age 65 to 69, 11: Age 70 to 74, 12: Age 75 to 79, 13: Age 80 or older, |
| _RACE | Race | Computed Race-Ethnicity grouping | 1: White only, non-Hispanic<br>2: Black only, non-Hispanic<br>3: American Indian or Alaskan Native only, Non-Hispanic 4: Asian only, non-Hispanic 5: Native Hawaiian or other Pacific Islander only, Non-Hispanic<br>6: Other race only, non-Hispanic 7: Multiracial, non-Hispanic 8: Hispanic | 1: White, 2: Black<br>3: American Indian<br>4: Asian, 5: Hawaiian<br>6: Other, 7: Multiracial<br>8: Hispanic |
| _SEX | Gender | Calculated sex variable | 1: Male, 2: Female | 0: Female, 1: Male |
| INCOME3 | Income Level | Income Level | 1: Less than $10,000, 2: Less than $15,000, 3: Less than $20,000, 4: Less than $25,000, 5: Less than $35,000,<br>6: Less than $50,000, 7: Less than $75,000,<br>8: Less than $100,000, 9: Less than $150,000,<br>10: Less than $200,000, 11: $200,000 or more,<br>77: N/A 99: Refused | 1: Less than $10,000, 2: Less than $15,000,<br>3: Less than $20,000, 4: Less than $25,000,<br>5: Less than $35,000, 6: Less than $50,000,<br>7: Less than $75,000, 8: Less than $100,000,<br>9: Less than $150,000, 10: Less than $200,000, 11: $200,000 or more, |
| EDUCA | Education level | What is the highest grade or year of school completed? | 1: Never attended school or only kindergarten,<br>2: Grades 1 through 8 (Elementary),<br>3: Grades 9 through 11 (Some high school),<br>4: Grade 12 or GED (High school graduate),<br>5: College 1 year to 3 years (Some college or technical school), 6: College 4 years or more (College graduate), | 0: Never attended school or only kindergarten,<br>1: Grades 1 through 8, 2: Grades 9 through 11,<br>3: Grade 12 or GED,<br>4: College 1 year to 3 years,<br>5: College 4 years or more (College graduate) |
| EMPLOY1 | Employment statues | Are you currently working? | 1: Employed for wages,<br>2: Self-employed,<br>3: Out of work for 1 year or more,<br>4: Out of work for < 1 year,<br>5: A homemaker, 6: A student<br>7: Retired, 8: Unable to work | 1: Employed for wages, 2: Self-employed, 3: Out of work for 1 year or more, 4: Out of work for < 1 year, 5: A homemaker, 6: A student, 7: Retired, 8: Unable to work |
| MARITAL | Marital status | Are you married? | 1: Married, 2: Divorced,<br>3: Widowed, 4: Separated, 5: Never married. 6: A member of an unmarried couple | 0: Never married, 1: Married, 2: Divorced,<br>3: Widowed, 4: Separated,<br>5: A member of an unmarried couple |
| RENTHOM1 | Housing Status | Do you own or rent your home? | 1: Own, 2: Rent<br>3: Other arrangement | 1: Own, 2: Rent,<br>3: Other arrangement |

**Table 10:** Adverse Childhood Experiences (ACEs) features in BRFSS 2019 and 2021.

| Feature Name Original | Feature Name(New) | Description | Value Type | Feature Map |
|---|---|---|---|---|
| ACEDIVRC | Parents Separation | Were parents Divorced/ Separated? | 1: Yes, 2: No, 8: Not married | 0: No, 1: Yes, 2: Not married |
| ACEPUNCH | Physically hurt | How often did parents beat each other up? | 1: Never, 2: Once<br>3: More than once | 0: No, 1: Once, 2: More |
| ACEDEPRS | Depressed Accompany | Live with anyone Depressed, Mentally Ill, Or Suicidal? | 1: Yes, 2: No | 0: No, 1: Yes |
| ACESWEAR | Verbally Insulted | How often did A Parent swear? | 1: Never, 2: Once<br>3: More than once | 0: No, 1: Once, 2: More |
| ACETOUCH | Touched Sexually | How often did anyone ever touch them sexually? | 1: Never, 2: Once 3: More than once | 0: No, 1: Once, 2: More |
| ACEHVSEX | Sexual harassment | How often was anyone ever forced to have sex? | 1: Never, 2: Once 3: More than once | 0: No, 1: Once, 2: More |
| ACEADNED | Basic Needs | Did an adult make sure basic needs were met? | 1: Never, 2: A little of the time, 3: Some of the time, 4: Most of the time, 5: All of the time. | 0: No, 1: A little of the time, 2: Some of the time,3: Most of the time, 4: All of the time. |
| ACEDRUGS | Drug Addicted F_member | Live with anyone who used illegal drugs or abused prescriptions? | 1: Yes, 2: No | 0: No, 1: Yes |

**Table 11:** This table demonstrates each feature with its minimum, maximum, mean, standard deviation, empty values and classification as categorical or numerical for BRFSS 2021 dataset.

| Index | Columns | Minimum | Maximum | Mean | Std | Values | Empty | Status |
|---|---|---|---|---|---|---|---|---|
| 1 | Parents Separation | 0 | 2 | 0.265 | 0.474 | 57285 | 381408 | Categorical |
| 2 | Physically Hurt | 1 | 3 | 1.262 | 0.647 | 56545 | 382148 | Categorical |
| 3 | Depressed Accompany | 0 | 1 | 0.180 | 0.384 | 57260 | 381433 | Categorical |
| 4 | Verbally Insulted | 1 | 3 | 1.599 | 0.885 | 56507 | 382186 | Categorical |
| 5 | Touched Sexually | 1 | 3 | 1.178 | 0.534 | 56573 | 382120 | Categorical |
| 6 | Sexual Harassment | 1 | 3 | 1.075 | 0.359 | 56593 | 382100 | Categorical |
| 7 | Basic Needs | 1 | 5 | 4.756 | 0.679 | 57112 | 381581 | Categorical |
| 8 | Drug Addicted F_member | 0 | 1 | 0.093 | 0.291 | 57419 | 381274 | Categorical |
| 9 | Age | 1 | 13 | 7.585 | 3.562 | 429086 | 9607 | Numerical |
| 10 | Race | 0 | 7 | 1.294 | 1.211 | 427703 | 10990 | Categorical |
| 11 | Gender | 1 | 2 | 1.535 | 0.498 | 438693 | 0 | Categorical |
| 12 | Income Level | 1 | 11 | 6.698 | 2.422 | 344280 | 94413 | Numerical |
| 13 | Education Level | 1 | 6 | 5.013 | 1.007 | 436215 | 2478 | Categorical |
| 14 | Employment Status | 0 | 1 | 0.520 | 0.499 | 430442 | 8251 | Categorical |
| 15 | Marital Status | 0 | 6 | 1.427 | 1.321 | 433705 | 4988 | Categorical |
| 16 | Smoking Status | 0 | 1 | 0.404 | 0.490 | 414232 | 24461 | Categorical |
| 17 | Alcohol Consumption | 0 | 30 | 4.800 | 8.073 | 408104 | 30589 | Categorical |
| 18 | Physical Activity | 0 | 1 | 0.755 | 0.429 | 437765 | 928 | Categorical |
| 19 | Regular Checkup | 0 | 4 | 1.349 | 0.776 | 432788 | 5905 | Categorical |
| 20 | Dietary Habits (Fruits) | 0 | 3 | 1.606 | 0.565 | 394742 | 43951 | Numerical |
| 21 | Dietary Habits (Vegetables) | 0 | 3 | 1.592 | 0.563 | 390165 | 48528 | Numerical |
| 22 | Housing Status | 1 | 3 | 1.337 | 0.566 | 434094 | 4599 | Categorical |
| 23 | DEAF | 0 | 1 | 0.088 | 0.284 | 422295 | 16398 | Categorical |
| 24 | BMI | 1 | 4 | 3.00 | 0.833 | 391841 | 46852 | Categorical |
| 25 | BLIND | 1 | 2 | 1.948 | 0.221 | 421342 | 17351 | Categorical |
| 26 | Difficulty Concentrating | 1 | 2 | 1.889 | 0.313 | 418713 | 19980 | Categorical |
| 27 | DIFFWALK | 1 | 2 | 1.838 | 0.367 | 418863 | 19830 | Categorical |
| 28 | Health Insurance | 1 | 2 | 1.055 | 0.228 | 421296 | 17397 | Categorical |
| 29 | Mental Health | 1 | 3 | 1.498 | 0.711 | 430776 | 7917 | Categorical |
| 30 | Physical Health | 1 | 3 | 1.448 | 0.696 | 429199 | 9494 | Categorical |
| 31 | Blood Pressure | 1 | 2 | 1.394 | 0.488 | 436781 | 1912 | Categorical |
| 32 | Cholesterol Awareness | 1 | 3 | 1.186 | 0.552 | 408911 | 29782 | Categorical |
| 33 | Kidney Disease | 0 | 1 | 0.039 | 0.193 | 436880 | 1813 | Categorical |

**Table 12: Baseline characteristics of the study population stratified by kidney disease status.**

| Features | No Kidney Disease | Kidney Disease | P-Value |
|---|---|---|---|

| Total (N=438693) | (n=419779) | (n=17101) | |
|---|---|---|---|
| **Age** | | | <0.001* |
| 18-24 | 25826 (6.2%) | 155 (0.9%) | |
| 25-29 | 21429 (5.1%) | 170 (1.0%) | |
| 30-34 | 25394 (6.0%) | 292 (1.7%) | |
| 35-39 | 28120 (6.7%) | 381 (2.2%) | |
| 40-44 | 28886 (6.9%) | 482 (2.8%) | |
| 45-49 | 27737 (6.6%) | 580 (3.4%) | |
| 50-54 | 33041 (7.9%) | 946 (5.5%) | |
| 55-59 | 36523 (8.7%) | 1348 (7.9%) | |
| 60-64 | 42264 (10.1%) | 1881 (11.0%) | |
| 65-69 | 43176 (10.3%) | 2386 (14.0%) | |
| 70-74 | 39895 (9.5%) | 2823 (16.5%) | |
| 75-79 | 27244 (6.5%) | 2330 (13.6%) | |
| 80+ | 31121 (7.4%) | 3085 (18.0%) | |
| **Gender** | | | <0.001* |
| Female | 224295 (53.4%) | 9662 (56.5%) | |
| Male | 195484 (46.6%) | 7439 (43.5%) | |
| **Race** | | | <0.001* |
| American Indian/Alaskan | 6884 (1.6%) | 359 (2.1%) | |
| Asian | 11148 (2.7%) | 221 (1.3%) | |
| Black | 31107 (7.4%) | 1535 (9.0%) | |
| Hispanic | 37056 (8.8%) | 1196 (7.0%) | |
| Multiracial | 8825 (2.1%) | 371 (2.2%) | |
| Native Hawaiian | 1906 (0.5%) | 73 (0.4%) | |
| Other | 3791 (0.9%) | 135 (0.8%) | |
| White | 308732 (73.5%) | 12820 (75.0%) | |
| **Education** | | | <0.001* |
| Elementary | 16098 (3.8%) | 929 (5.4%) | |
| High School Grad | 114452 (27.3%) | 5186 (30.3%) | |
| No School | 7665 (1.8%) | 502 (2.9%) | |
| Some College | 172369 (41.1%) | 5706 (33.4%) | |
| Some High School | 106392 (25.3%) | 4669 (27.3%) | |
| **Employment** | | | <0.001* |
| Employed | 182802 (43.5%) | 2912 (17.0%) | |
| Homemaker | 16905 (4.0%) | 613 (3.6%) | |
| Out of work (<1yr) | 9452 (2.3%) | 231 (1.4%) | |
| Out of work (>1yr) | 10906 (2.6%) | 445 (2.6%) | |
| Retired | 121706 (29.0%) | 9094 (53.2%) | |
| Self-employed | 36885 (8.8%) | 787 (4.6%) | |
| Student | 10559 (2.5%) | 81 (0.5%) | |
| Unable to work | 22741 (5.4%) | 2708 (15.8%) | |
| **Parents Separation** | | | <0.001* |
| 0.0 | 40862 (9.7%) | 1918 (11.2%) | |
| 1.0 | 12975 (3.1%) | 450 (2.6%) | |
| 2.0 | 816 (0.2%) | 39 (0.2%) | |
| **Verbally Insulted** | | | 0.238 |
| 0.0 | 36224 (8.6%) | 1636 (9.6%) | |
| 1.0 | 3022 (0.7%) | 121 (0.7%) | |
| 2.0 | 14663 (3.5%) | 622 (3.6%) | |

**Table 13:** Activity- and diet-related features from BRFSS 2019 and 2021.

| Feature Name Raw | Feature Name (New) | Description | Raw categories | Feature category map |
|---|---|---|---|---|
| SMOKE100 | Smoking status | Smoked at Least 100 Cigarettes | 1: Yes, 2: No | 0: No, 1: Yes |
| ALCDAY5 | Alcohol consumption | Days in past 30 had alcoholic beverage | 101 - 107: Days per week, 201 - 230: Days in past 30 days, 777: N/A, 888: No drinks in past 30 days, 999: N/A, | 0: No drinks in past 30 days, 1: 1 day, 2: 2days, 3: 3days, 4: 4days, 5: 5days, 6: 6 days, 7:7 days, 8: 8days, 9: 9days, 10:10 days, 11: days, 12: 12days, 13:13 days, 14: 14 days, 15:15 days, 16: 16 days, 17:17 days, 18: 18 days, 19:19 days, 20:20 days, 21: days, 22: 22 days, 23: 23 days, 24:24 days, 25: 25days, 26: 26 days, 27:27 days, 28: 28 days, 29: 29 days, 30: drinks in past 30 days |
| EXERANY2 | Physical activity | Exercise in Past 30 Days | 1: Yes, 2: No | 0: No, 1: Yes |
| CHECKUP1 | Regular Checkup | Length of time since last routine checkup | 1: Within past year (anytime < 12 months ago, 2: Within past 2 years (1 year but < 2 years ago), 3: Within past 5 years (2 years but < 5 years ago), 4: 5 or more years ago, 8: Never | 0: Never<br>1: Within past year<br>2: Within past 2 years<br>3: Within past 5 years<br>4: 5 or more years ago |
| FRUIT2 | Dietary Habits (Fruits) | How many times did you eat fruit? | 101 - 199: Days , 201 - 299: Weeks<br>300: Less than once a month,<br>301 - 399: Month / Year ,<br>555: Never, 777: N/A, 999: N/A | 0: Never, [300]<br>1: 1% [101,201,301],<br>2: 2-98% [102-198, 202-298, 302-398],<br>3: 99%[ 99,199, 299], |
| VEGETAB2 | Dietary Habits (Vegetables) | How many times did you eat OTHER vegetables? | 101 - 199: Days<br>201 - 299: Weeks<br>300: Less than once a month<br>301 - 399: Month / Year<br>555: Never, 777: N/A, 999: N/A | 0: Never,[300]<br>1: 1% [101,201,301],<br>2: 2-98% [102-198, 202-298, 302-398],<br>3: 99%[ 99,199, 299], |

**Table 14.** Architectural details and hyperparameter settings for all predictive models.

| Model Category | Model Name | Hyperparameters & Architecture Details |
|---|---|---|
| | **Logistic Regression (LR)** | max_iter: 1000, random_state: 42 |
| | **Decision Tree (DT)** | criterion: 'gini' (default), random_state: 42 |
| | **Random Forest (RF)** | n_estimators: 100, random_state: 42 |
| **Traditional ML Models** | **XGBoost** | eval_metric: 'logloss', use_label_encoder: False, random_state: 42 |
| | **AdaBoost** | n_estimators: 50 (default), random_state: 42 |
| | **Gradient Boosting** | n_estimators: 100 (default), random_state: 42 |
| | **LightGBM** | verbose: -1, random_state: 42 |
| | **ANN 1-Layer** | **Layers:** Dense(64, ReLU) → Dense(1, Sigmoid), Optimizer: Adam, Loss: Binary Crossentropy<br>**Training:** Epochs: 50, Batch Size: 32<br>Callback: Early Stopping (monitor: 'loss', patience: 5) |
| **Deep Learning** | **ANN 2-Layer** | **Layers:** Dense(128, ReLU) → Dense(64, ReLU) → Dense(1, Sigmoid), Optimizer: Adam, Loss: Binary Crossentropy<br>**Training:** Epochs: 50, Batch Size: 32 |
| | | Callback: Early Stopping (monitor: 'loss', patience: 5) |
| | **ANN+Attention** | **Layers:** Self-Attention (Softmax) × Input → Dense(64, ReLU) → Dense(32, ReLU) → Dense(1, Sigmoid), Optimizer: Adam, Loss: Binary Crossentropy<br>**Training:** Epochs: 50, Batch Size: 32<br>Callback: Early Stopping (monitor: 'loss', patience: 5) |
| **Stacked Architecture** | **Ensemble Stacking** | **Base Estimators:**<br>**1. Logistic Regression** (max_iter: 1000)<br>**2. Random Forest** (n_estimators: 100)<br>**3. XGBoost (eval_metric: 'logloss')**<br>**Meta-Learner (Final Estimator):**<br>**MLPClassifier** (hidden_layer_sizes: (32,), max_iter: 500, random_state: 42) |

**Supplementary Table 1:** This table BRFSS 2019, clearly displays the minimum, maximum, mean, standard deviation (std) and missing or empty values for each feature we worked with, along with their classification as categorical or numerical.

| Index | Columns | Minimum | Maximum | Mean | Std | Values | Empty | Status |
|---|---|---|---|---|---|---|---|---|
| 1 | Parents Separation | 0 | 2 | 0.27 | 0.47 | 106015 | 312253 | Categorical |
| 2 | Physically hurt | 1 | 3 | 1.27 | 0.66 | 104654 | 313614 | Categorical |
| 3 | Depressed Accompany | 0 | 1 | 0.16 | 0.37 | 105996 | 312272 | Categorical |
| 4 | Verbally Insulted | 1 | 3 | 1.58 | 0.87 | 104632 | 313636 | Categorical |
| 5 | Touched Sexually | 1 | 3 | 1.18 | 0.54 | 105005 | 313263 | Categorical |
| 6 | Sexual harassment | 1 | 3 | 1.07 | 0.36 | 105082 | 313186 | Categorical |
| 7 | Basic Needs | 1 | 5 | 4.756 | 0.679 | 57112 | 381581 | Categorical |
| 8 | Drug Addicted F_member | 0 | 1 | 0.09 | 0.28 | 106220 | 312048 | Categorical |
| 9 | Age | 1 | 13 | 7.72 | 3.57 | 411580 | 6688 | Numerical |
| 10 | Race | 1 | 8 | 2.00 | 2.20 | 409333 | 8935 | Categorical |
| 11 | Gender | 0 | 1 | 0.45 | 0.49 | 418268 | 0 | Categorical |
| 12 | Income Level | 1 | 8 | 5.94 | 2.13 | 338487 | 79781 | Numerical |
| 13 | Education level | 1 | 6 | 4.93 | 1.03 | 416433 | 1835 | Categorical |
| 14 | Employment statues | 1 | 8 | 3.90 | 2.86 | 411761 | 6507 | Categorical |
| 15 | Marital status | 1 | 6 | 2.30 | 1.65 | 414784 | 3484 | Categorical |
| 16 | Smoking status | 0 | 1 | 0.42 | 0.49 | 399729 | 18539 | Categorical |
| 17 | Alcohol consumption | 0 | 30 | 4.71 | 8.05 | 395114 | 23154 | Categorical |
| 18 | Physical activity | 0 | 1 | 0.72 | 0.44 | 396261 | 22007 | Categorical |
| 19 | Regular Checkup | 0 | 4 | 1.31 | 0.77 | 413448 | 4820 | Categorical |
| 20 | Dietary Habits (Fruits) | 0 | 2 | 1.59 | 0.57 | 379248 | 39020 | Numerical |
| 21 | Dietary Habits (Vegetables) | 0 | 2 | 1.60 | 0.56 | 374899 | 43369 | Numerical |
| 22 | Housing Status | 1 | 3 | 1.34 | 0.57 | 414877 | 3391 | Categorical |
| 23 | DEAF | 0 | 1 | 0.09 | 0.29 | 405410 | 12858 | Categorical |
| 24 | BMI | 1 | 4 | 2.98 | 0.83 | 382065 | 36203 | Categorical |
| 25 | BLIND | 1 | 9 | 1.96 | 0.39 | 406205 | 12063 | Categorical |
| 26 | Difficulty concentrating | 0 | 1 | 0.11 | 0.31 | 402589 | 15679 | Categorical |
| 27 | DIFFWALK | 0 | 1 | 0.17 | 0.38 | 402917 | 15351 | Categorical |
| 28 | Health Insurance | 0 | 1 | 0.91 | 0.27 | 416193 | 2075 | Categorical |
| 29 | Mental Health | 1 | 3 | 1.46 | 0.70 | 409489 | 8779 | Categorical |
| 30 | Physical Health | 1 | 3 | 1.52 | 0.73 | 407859 | 10409 | Categorical |
| 31 | Blood pressure | 1 | 2 | 1.40 | 0.49 | 416696 | 1572 | Categorical |
| 32 | Cholesterol Awareness | 1 | 3 | 1.15 | 0.50 | 393720 | 24548 | Categorical |
| 33 | Kidney Disease | 0 | 1 | 0.03 | 0.19 | 416661 | 1607 | Categorical |

**Supplementary Table 2:** This table NHIS 2020, clearly displays the minimum, maximum, mean, standard deviation (std) and missing or empty values for each feature we worked with, along with their classification as categorical or numerical.

| Index | Columns | Minimum | Maximum | Mean | Std | Values | Empty | Status |
|---|---|---|---|---|---|---|---|---|
| 1 | Age | 1 | 14 | 7.32 | 3.63 | 31498 | 70 | Numerical |
| 2 | Race | 1 | 7 | 2.18 | 0.91 | 31568 | 0 | Categorical |
| 3 | Gender | 0 | 1 | 0.46 | 0.49 | 31566 | 2 | Categorical |
| 4 | Income Level | 1 | 14 | 10.03 | 3.93 | 31568 | 0 | Numerical |
| 5 | Education level | 0 | 11 | 6.06 | 2.45 | 31419 | 149 | Categorical |
| 6 | Employment statues | 1 | 6 | 1.85 | 1.45 | 19388 | 12180 | Categorical |
| 7 | Marital status | 0 | 2 | 0.60 | 0.60 | 30590 | 978 | Categorical |
| 8 | Smoking status | 0 | 1 | 0.37 | 0.48 | 30986 | 582 | Categorical |
| 9 | Regular Checkup | 0 | 6 | 1.95 | 1.52 | 6162 | 25406 | Categorical |
| 10 | Housing Status | 1 | 3 | 1.32 | 0.51 | 30385 | 1183 | Categorical |
| 11 | DEAF | 0 | 3 | 0.18 | 0.43 | 31558 | 10 | Categorical |
| 12 | BMI | 1 | 4 | 2.97 | 0.83 | 30839 | 729 | Categorical |
| 13 | BLIND | 0 | 3 | 0.19 | 0.43 | 31557 | 11 | Categorical |
| 14 | Difficulty concentrating | 0 | 3 | 0.20 | 0.45 | 31551 | 17 | Categorical |
| 15 | Difficulty Walking | 0 | 3 | 0.29 | 0.61 | 31558 | 10 | Categorical |
| 16 | Health Insurance | 0 | 1 | 0.93 | 0.24 | 31532 | 36 | Categorical |
| 17 | Mental Health | 0 | 1 | 0.10 | 0.30 | 31157 | 411 | Categorical |
| 18 | Blood pressure | 0 | 1 | 0.83 | 0.37 | 11481 | 20087 | Categorical |
| 19 | Cholesterol Awareness | 0 | 1 | 0.31 | 0.46 | 31460 | 108 | Categorical |
| 20 | Kidney Disease | 0 | 1 | 0.03 | 0.18 | 17690 | 13878 | Categorical |

**Supplementary Table 3:** This table, NHIS 2021, clearly displays the minimum, maximum, mean, standard deviation (std) and missing or empty values for each feature we worked with, along with their classification as categorical or numerical.

| Index | Columns | Minimum | Maximum | Mean | Std | Values | Empty | Status |
|---|---|---|---|---|---|---|---|---|
| 1 | Age | 1 | 14 | 7.12 | 3.67 | 29396 | 86 | Numerical |
| 2 | Race | 1 | 7 | 2.20 | 0.96 | 29482 | 0 | Categorical |
| 3 | Gender | 0 | 1 | 0.45 | 0.49 | 29480 | 2 | Categorical |
| 4 | Income Level | 1 | 14 | 9.84 | 4.01 | 29482 | 0 | Numerical |
| 5 | Education level | 1 | 10 | 6.51 | 2.34 | 29397 | 85 | Categorical |
| 6 | Employment status | 1 | 8 | 3.55 | 1.40 | 12178 | 17304 | Categorical |
| 7 | Marital status | 0 | 2 | 0.59 | 0.60 | 28537 | 945 | Categorical |
| 8 | Smoking status | 0 | 1 | 0.37 | 0.48 | 28596 | 886 | Categorical |
| 9 | Regular Checkup | 0 | 6 | 2.03 | 1.52 | 5749 | 23733 | Categorical |
| 10 | Housing Status | 1 | 3 | 1.33 | 0.51 | 28254 | 1228 | Categorical |
| 11 | DEAF | 0 | 3 | 0.18 | 0.43 | 29467 | 15 | Categorical |
| 12 | BMI | 1 | 4 | 2.97 | 0.83 | 28749 | 733 | Categorical |
| 13 | BLIND | 0 | 3 | 0.19 | 0.44 | 29464 | 18 | Categorical |
| 14 | Difficulty concentrating | 0 | 3 | 0.24 | 0.49 | 29469 | 13 | Categorical |
| 15 | Difficulty Walking | 0 | 3 | 0.28 | 0.61 | 29470 | 12 | Categorical |
| 16 | Health Insurance | 0 | 1 | 0.93 | 0.24 | 29443 | 39 | Categorical |
| 17 | Mental Health | 0 | 1 | 0.11 | 0.31 | 28878 | 604 | Categorical |
| 18 | Blood pressure | 0 | 1 | 0.82 | 0.38 | 10646 | 18836 | Categorical |
| 19 | Cholesterol Awareness | 0 | 1 | 0.30 | 0.46 | 29391 | 91 | Categorical |
| 20 | Kidney Disease | 0 | 1 | 0.03 | 0.18 | 29450 | 32 | Categorical |

## NHIS 2020 Feature map

**Supplementary Table 4:** This table NHIS 2020 maps demographic characteristics, with clear descriptions linking the original characteristic names to the new names.

| Feature Name Original | Feature Name (New) | Description | Value Type | Feature Map (NHIS 2020) |
|---|---|---|---|---|
| AGEP_A | Age | Reported age in five-year age categories calculated variable | 18-84: 18-84 years<br>85: 85+ years | 1: Age 18 to 24,2: Age 25 to 29,<br>3: Age 30 to 34, 4: Age 35 to 39,<br>5: Age 40 to 44, 6: Age 45 to 49,<br>7: Age 50 to 54, 8: Age 55 to 59,<br>9: Age 60 to 64, 10: Age 65 to 69,<br>11: Age 70 to 74, 12: Age 75 to 79,<br>13: Age 80 to 84<br>14: Age 85+ |
| HISPALLP_A | Race | Computed Race-Ethnicity grouping | 1: Hispanic<br>2: Non-Hispanic White only<br>3: Non-Hispanic Black/African American only<br>4: Non-Hispanic Asian only<br>5: Non-Hispanic AIAN only<br>6: Non-Hispanic AIAN and any other group<br>7: Other single and multiple races | 1:Hispanic<br>2: Non-Hispanic White only<br>3: Non-Hispanic Black/African American only<br>4:Non-Hispanic Asian only<br>5:Non-Hispanic AIAN only<br>6:Non-Hispanic AIAN and any other group<br>7:Other single and multiple races |
| SEX_A | Gender | Calculated sex variable | 1: Male<br>2: Female | 0: Female, 1: Male |
| RATCAT_A | Income Level | Income Level | 1: 0.00 - 0.49<br>2: 0.50 - 0.74<br>3: 0.75 - 0.99<br>4: 1.00 - 1.24<br>5: 1.25 - 1.49<br>6: 1.50 - 1.74<br>7: 1.75 - 1.99<br>8: 2.00 - 2.49<br>9: 2.50 - 2.99<br>10: 3.00 - 3.49<br>11: 3.50 - 3.99<br>12: 4.00 - 4.49<br>13:4.50 - 4.99<br>14:5.00 or greater | 1: 0.00 - 0.49<br>2: 0.50 - 0.74<br>3: 0.75 - 0.99<br>4: 1.00 - 1.24<br>5: 1.25 - 1.49<br>6: 1.50 - 1.74<br>7: 1.75 - 1.99<br>8: 2.00 - 2.49<br>9: 2.50 - 2.99<br>10: 3.00 - 3.49<br>11: 3.50 - 3.99<br>12: 4.00 - 4.49<br>13:4.50 - 4.99<br>14: 5.00 or greater |
| EDUC_A | Education level | What is the highest grade or year of school completed? | 0: Never attended/kindergarten only<br>1: Grade<br>2: 12th grade, no diploma<br>3: GED or equivalent<br>4: High School Graduate<br>5: Some college, no degree 3925<br>6: Associate degree: occupational, technical, or vocational program<br>7: Associate degree: academic program<br>8: Bachelor's degree (Example: BA, AB, BS, BBA)<br>9: Master's degree (Example: MA, MS, MEng, | 0: Never attended/kindergarten only<br>1: Grade<br>2: 12th grade, no diploma<br>3: GED or equivalent<br>4: High School Graduate<br>5: Some college, no degree 3925<br>6: Associate degree: occupational, technical, or vocational program<br>7: Associate degree: academic program<br>8: Bachelor's degree (Example: BA, AB, BS, BBA)<br>9: Master's degree (Example: MA, MS, |

| | | | MEd, MBA)<br>10: Professional School or Doctoral degree (Example: MD, DDS, DVM, JD, PhD, EdD) | MEng, MEd, MBA)<br>10: Professional School or Doctoral degree (Example: MD, DDS, DVM, JD, PhD, EdD) |
|---|---|---|---|---|
| EMDWRKCAT_A | Employment statues | Are you currently working? | 1: Employee of a PRIVATE company for wages<br>2: A FEDERAL government employee<br>3: A STATE government employee<br>4: A LOCAL government employee<br>5: Self-employed in OWN business, professional practice or farm<br>6: Working WITHOUT PAY in a family-owned business or farm | 1: Employee of a PRIVATE company for wages<br>2: A FEDERAL government employee<br>3: A STATE government employee<br>4: A LOCAL government employee<br>5: Self-employed in OWN business, professional practice or farm<br>6: Working WITHOUT PAY in a family-owned business or farm |
| MARITAL_A | Marital status | Are you married? | 1: Married<br>2: Living with a partner together as an unmarried couple<br>3: Neither | 0: Neither, 1: Married, 2: Living with a partner together as an unmarried couple. |
| HOUTENURE_A | Housing Status | Do you own or rent your home? | 1: Owned or being bought,<br>2: Rented,<br>3: Other arrangement | 1: Owned or being bought,<br>2: Rented,<br>3: Other arrangement |

**Supplementary Table 5:** This table (NHIS 2020) outlines the activity-related features, with clear descriptions mapping the original feature names to the new names.

| Feature Name Raw | Feature Name (New) | Description | Raw categories | Feature Map (NHIS 2020) |
|---|---|---|---|---|
| SMKEV_A | Smoking status | Smoked at Least 100 Cigarettes | 1: Yes, 2: No | 0: No, 1: Yes |
| WELLVIS_A | Regular Checkup | Length of time since last routine checkup | 0: Never<br>1: Within the past year<br>2: Within the last 2 years<br>3: Within the last 3 years<br>4: Within the last 5 years<br>5: Within the last 10 years<br>6: 10 years ago or more | 0: Never<br>1: Within the past year<br>2: Within the last 2 years<br>3: Within the last 3 years<br>4: Within the last 5 years<br>5: Within the last 10 years<br>6: 10 years ago or more |

**Supplementary Table 6:** This table (NHIS 2020) is about disability and health, with original feature names having been given new names.

| Feature Name Original | Feature Name (New) | Description | Value Type | Feature Map (NHIS 2020) |
|---|---|---|---|---|
| HEARINGDF_A | DEAF | Are you deaf or do you have serious difficulty hearing? | 1: No<br>2: Some difficulty<br>3: A lot of difficulty | 0: No,<br>1: Some difficulty<br>2: A lot of difficulty |

| | | | 4: Cannot do at all | 3: Cannot do at all |
|---|---|---|---|---|
| BMICAT_A | BMI | Computed body mass index categories | 1: Underweight, 2: Healthy weight<br>3: Overweight, 4: Obese | 1: Underweight, 2: Healthy weight<br>3: Overweight, 4: Obese |
| VISIONDF_A | BLIND | Blind or Difficulty seeing | 1: No<br>2: Some difficulty<br>3: A lot of difficulty<br>4: Cannot do at all | 0: No,<br>1: Some difficulty<br>2: A lot of difficulty<br>3: Cannot do at all |
| COGMEMDFF_A | Difficulty concentrating | Difficulty Concentrating or Remembering | 1: No<br>2: Some difficulty<br>3: A lot of difficulty<br>4: Cannot do at all | 0: No,<br>1: Some difficulty<br>2: A lot of difficulty<br>3: Cannot do at all |
| DIFF_A | Difficulty Walking | Difficulty Walking or Climbing Stairs | 1: No<br>2: Some difficulty<br>3: A lot of difficulty<br>4: Cannot do at all | 0: No,<br>1: Some difficulty<br>2: A lot of difficulty<br>3: Cannot do at all |
| HICOV_A | Health Insurance | Have any health insurance? | 1: Yes, 2: No | 0: No, 1: Yes |
| MHTHRPY_A | Mental Health | Computed Mental Health Status | 1: Yes, 2: No | 0: No, 1: Yes |
| HYPMED_A | Blood pressure | Have High Blood Pressure? | 1: Yes, 2: No | 0: No, 1: Yes |
| CHLEV_A | Cholesterol Awareness | Have Cholesterol Checked? | 1: Yes, 2: No | 0: No, 1: Yes |
| KIDWEAKEV_A | Kidney Disease | Have kidney disease? | 1: Yes, 2: No | 0: No, 1: Yes |

**Supplementary Table 7:** Data distribution of the BRFSS 2021 training set after resampling.

| Filling method | Sampling | Class 0 # | Class 1 # | Class 0 % | Class 1 % |
|---|---|---|---|---|---|
| XGBoost | No | 252,871 | 10,344 | 96.07 | 3.93 |
| | Oversampling | 252,871 | 252,871 | 50.00 | 50.00 |
| | Undersampling | 20,688 | 20,688 | 50.00 | 50.00 |
| | SMOTE | 252,871 | 252,871 | 50.00 | 50.00 |
| | AdaSyn | 252,871 | 250,368 | 50.25 | 49.75 |
| KNN | No | 252,880 | 10,335 | 96.07 | 3.93 |
| | Oversampling | 252,880 | 252,880 | 50.00 | 50.00 |
| | Undersampling | 10,335 | 10,335 | 50.00 | 50.00 |
| | SMOTE | 252,880 | 252,880 | 50.00 | 50.00 |
| | AdaSyn | 252,880 | 254,408 | 49.85 | 50.15 |
| GAN | No | 252,896 | 10,319 | 96.08 | 3.92 |
| | Oversampling | 252,896 | 252,896 | 50.00 | 50.00 |
| | Undersampling | 10,319 | 10,319 | 50.00 | 50.00 |
| | SMOTE | 252,896 | 252,896 | 50.00 | 50.00 |
| | AdaSyn | 252,896 | 250,516 | 50.24 | 49.76 |
| Diffusion | No | 252,896 | 10,319 | 96.08 | 3.92 |
| | Oversampling | 252,896 | 252,896 | 50.00 | 50.00 |
| | Undersampling | 10,319 | 10,319 | 50.00 | 50.00 |
| | SMOTE | 252,896 | 252,896 | 50.00 | 50.00 |
| | AdaSyn | 252,896 | 254,754 | 49.82 | 50.18 |
| Autoencoder-5 | No | 252,892 | 10,323 | 96.08 | 3.92 |
| | Oversampling | 252,892 | 10,323 | 96.08 | 3.92 |
| | Undersampling | 10,323 | 10,323 | 50.00 | 50.00 |
| | SMOTE | 252,892 | 252,892 | 50.00 | 50.00 |
| | AdaSyn | 252,892 | 254,864 | 49.81 | 50.19 |
| Autoencoder-15 | No | 252,888 | 10,327 | 96.08 | 3.92 |
| | Oversampling | 252,888 | 252,888 | 50.00 | 50.00 |
| | Undersampling | 10,327 | 10,327 | 50.00 | 50.00 |
| | SMOTE | 252,888 | 252,888 | 50.00 | 50.00 |
| | AdaSyn | 252,888 | 255,037 | 49.79 | 50.21 |
| Matrix Factorization | No | 251,786 | 11,429 | 95.66 | 4.34 |
| | Oversampling | 251,786 | 251,786 | 50.00 | 50.00 |
| | Undersampling | 11,429 | 11,429 | 50.00 | 50.00 |
| | SMOTE | 251,786 | 251,786 | 50.00 | 50.00 |
| | AdaSyn | 251,786 | 249,933 | 50.18 | 49.82 |

**Supplementary Table 8:** Data distribution of the BRFSS 2021 training set after resampling.

| Filling method | Sampling | Class 0 # | Class 1 # | Class 0 % | Class 1 % |
|---|---|---|---|---|---|
| Forward-Backward | No | 252,839 | 10,376 | 96.06 | 3.94 |
| | Oversampling | 252,839 | 252,839 | 50.00 | 50.00 |
| | Undersampling | 10,376 | 10,376 | 50.00 | 50.00 |
| | SMOTE | 252,839 | 252,839 | 50.00 | 50.00 |
| | AdaSyn | 252,839 | 254,578 | 49.83 | 50.17 |
| Mean | No | 252,896 | 10,319 | 96.08 | 3.92 |
| | Oversampling | 252,896 | 252,896 | 50.00 | 50.00 |
| | Undersampling | 10,319 | 10,319 | 50.00 | 50.00 |
| | SMOTE | 252,896 | 252,896 | 50.00 | 50.00 |
| | AdaSyn | 252,896 | 254,491 | 49.84 | 50.16 |
| Drop Nan | No | 19,362 | 826 | 95.91 | 4.09 |
| | Oversampling | 19,362 | 19,362 | 50.00 | 50.00 |
| | Undersampling | 826 | 826 | 50.00 | 50.00 |
| | SMOTE | 19,362 | 19,362 | 50.00 | 50.00 |
| | AdaSyn | 19,362 | 19,731 | 49.53 | 50.47 |

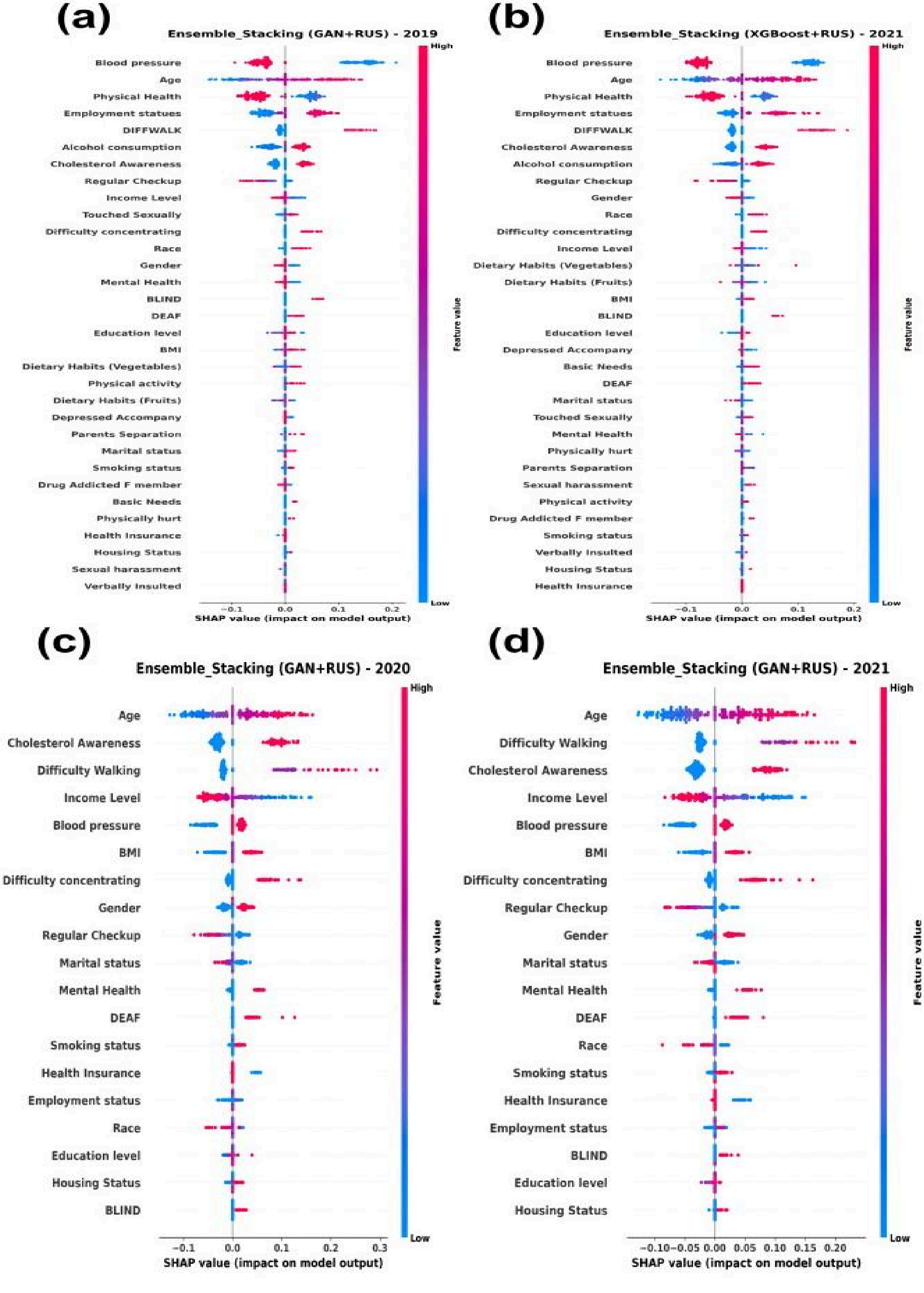

(a)
Ensemble_Stacking (GAN+RUS) - 2019
Blood pressure
Age
Physical Health
Employment statues
DIFFWALK
Alcohol consumption
Cholesterol Awareness
Regular Checkup
Income Level
Touched Sexually
Difficulty concentrating
Race
Gender
Mental Health
BLIND
DEAF
Education level
BMI
Dietary Habits (Vegetables)
Physical activity
Dietary Habits (Fruits)
Depressed Accompany
Parents Separation
Marital status
Smoking status
Drug Addicted F member
Basic Needs
Physically hurt
Health Insurance
Housing Status
Sexual harassment
Verbally Insulted
SHAP value (impact on model output)
High
Low
Feature value
(b)
Ensemble_Stacking (XGBoost+RUS) - 2021
Blood pressure
Age
Physical Health
Employment statues
DIFFWALK
Cholesterol Awareness
Alcohol consumption
Regular Checkup
Gender
Race
Difficulty concentrating
Income Level
Dietary Habits (Vegetables)
Dietary Habits (Fruits)
BMI
BLIND
Education level
Depressed Accompany
Basic Needs
DEAF
Marital status
Touched Sexually
Mental Health
Physically hurt
Parents Separation
Sexual harassment
Physical activity
Drug Addicted F member
Smoking status
Verbally Insulted
Housing Status
Health Insurance
SHAP value (impact on model output)
High
Low
Feature value
(c)
Ensemble_Stacking (GAN+RUS) - 2020
Age
Cholesterol Awareness
Difficulty Walking
Income Level
Blood pressure
BMI
Difficulty concentrating
Gender
Regular Checkup
Marital status
Mental Health
DEAF
Smoking status
Health Insurance
Employment status
Race
Education level
Housing Status
BLIND
−0.1 0.0 0.1 0.2 0.3
SHAP value (impact on model output)
High
Low
Feature value
(d)
Ensemble_Stacking (GAN+RUS) - 2021
Age
Difficulty Walking
Cholesterol Awareness
Income Level
Blood pressure
BMI
Difficulty concentrating
Regular Checkup
Gender
Marital status
Mental Health
DEAF
Race
Smoking status
Health Insurance
Employment status
BLIND
Education level
Housing Status
−0.10 −0.05 0.00 0.05 0.10 0.15 0.20
SHAP value (impact on model output)
High
Low
Feature value

**Supplementary Figure 1:** Beeswarm plots show the rank of patient features by their contributions to the CKD classifications. These plots show which patient details have the highest average impact on the model's decisions across four datasets: (a) BRFSS 2019, (b) BRFSS 2021, (c) NHIS 2020, and (d) NHIS 2021. Features are ranked in descending order by their average overall contribution. Red means a high value and blue means a low value. Dots placed to the right of the center line indicate that those specific features are contributing more to the model's predicted risk of CKD.

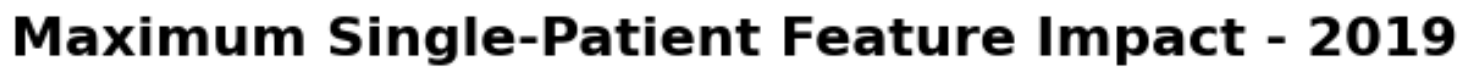

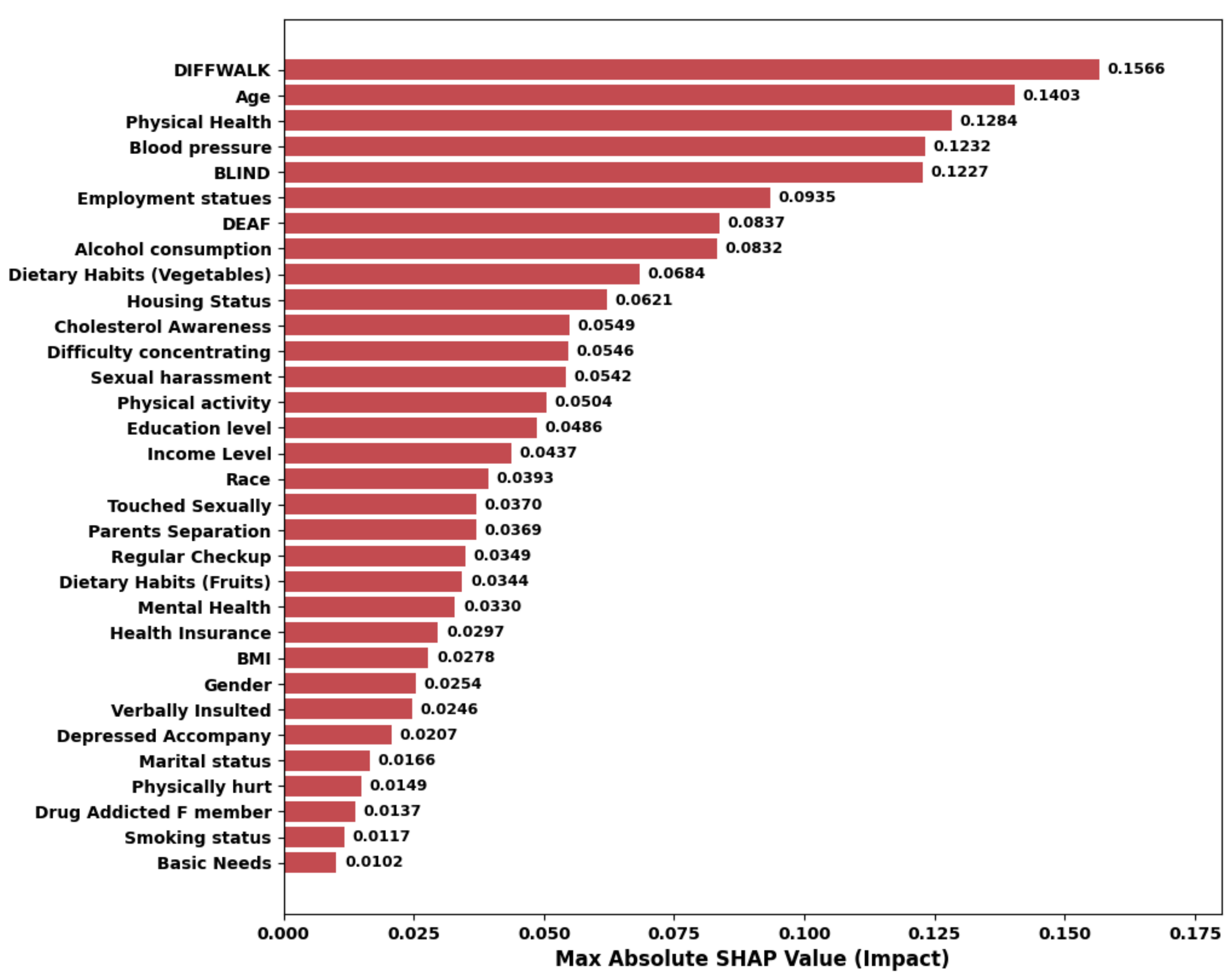

**Supplementary Figure 2:** Feature importance plots based on the maximum absolute SHAP values (max(|SHAP value|)) from the BRFSS 2019 dataset. It display feature importance using the Random Under Sampling (RUS) technique. The X-axis represents the magnitude of each feature's contribution to the model's prediction, while the Y-axis lists the input features.

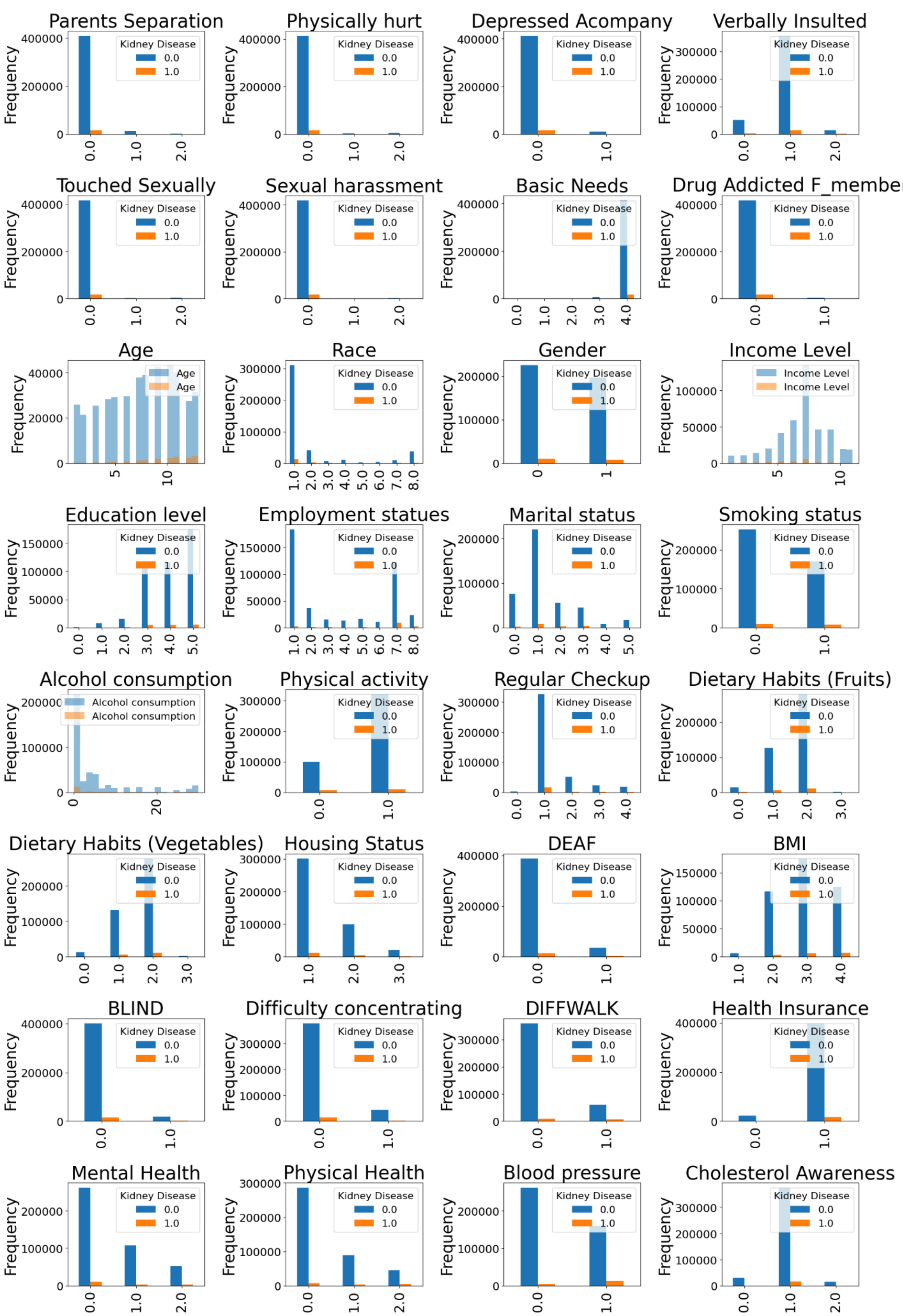

**Supplementary Figure 3:** Data Distribution Plot where Filling method was Diffusion

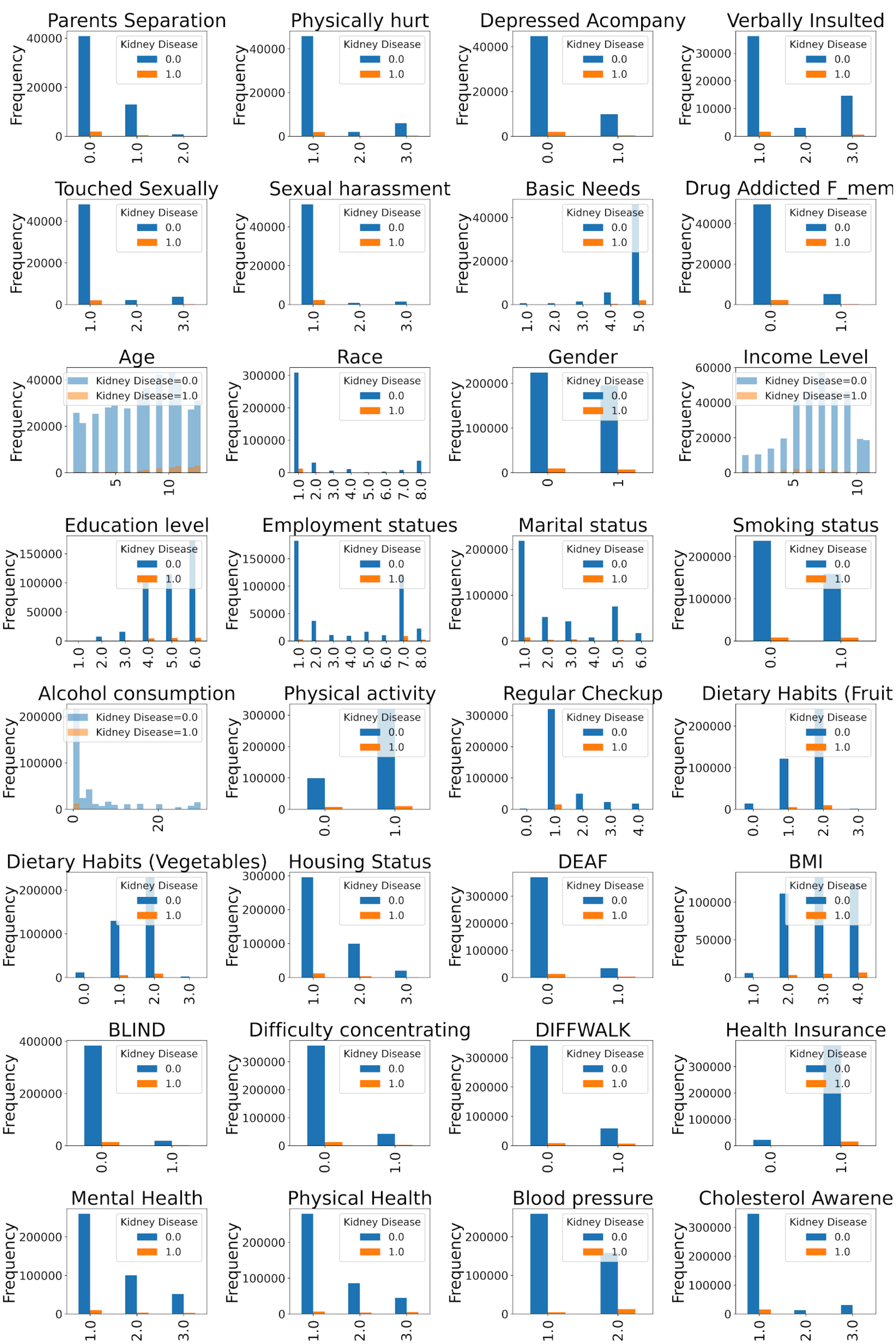

**Supplementary Figure 4:** Data Distribution Plot without any filling method.

The distribution plot demonstrates the distribution of data in terms of positive (1) and negative (0) cases of kidney disease. Negative cases are represented by blue, and positive cases are denoted by orange. Each subplot represents an individual column or feature of the dataset. The Y-axis of each subplot shows the number of rows or amount of data available, and the X-axis represents the categories of data available in a feature.

The plot shows the data distribution of 32 features, which includes demographic features (such as gender, age, race, marital status, etc.). There are features indicating adverse childhood experience (such as basic needs, drug-addicted family members, etc.) and some behavioral factors (such as alcohol consumption, smoking, physical activity, etc.).

**Discussion:**

Firstly, there are 2 types of factors that are regulated by the brain Neurotransmitter and Hormones Either way both of them can alter the function and biochemical response of the Kidney/Renal system. Norepinephrine, epinephrine, dopamine, serotonin, GABA, glutamate, and glycine are neurotransmitters that affect the renal system. Dopamine: Dopamine directly inhibits renal sodium reabsorption and decreases activity of the renin-angiotensin system at multiple steps, helping to maintain normal blood pressure. GRK4 variants decrease dopamine receptor activity and increase AT1R activity, with a net effect of increased sodium reabsorption, which predisposes to hypertension. ACE angiotensin-converting enzyme, AT1 angiotensin II type 1, GRK4 G-protein receptor kinase 4. So excess of Dopamine can cause hypertension[110]. Norepinephrine: Stress and the Fight or Flight Response: when an individual experience stress, the brain perceives a threat or danger, and the hypothalamus sends signals to the adrenal glands and the nervous system. This triggers the release of norepinephrine (noradrenaline) and epinephrine (adrenaline) from the adrenal glands. Serotonin: Mitochondrial Dysfunction Serotonin receptors can influence mitochondrial function in the kidneys, and altered serotonin levels may contribute to mitochondrial dysfunction, a factor in CKD progression[111].

The brain controls hormone and neurotransmitter release primarily through the hypothalamus and pituitary gland, working together as a central command center. Parental separation has a significant role in mental health deterioration and upbringing as well as personality buildup. Thus it affects the kidney system. Parental separation is a major cause of child mental health[112]. Hypertension can damage the kidneys, leading to reduced kidney function. Reduced kidney function can lead to fluid retention and increased blood volume, further raising blood pressure. This cycle can continue, leading to a worsening of both hypertension and CKD[113]. CKD is more common in older adults where 34% in those over 65 years of age vs 12% in those aged 45–64 years, because Glomerular filtration rate decreases over aging. With aging there are likely to be other factors that will contribute to CKD such as diabetes, hypertension, and heart diseases which are also associated with aging. But there is not any direct linear relation of aging to develop CKD[114]. Sex hormones, like estrogen and testosterone, are thought to play a role in the differences observed in CKD. Estrogen can have protective effects on kidney function, potentially contributing to the slower progression of CKD in women. Testosterone, on the other hand, can have a detrimental effect on kidney function, potentially contributing to the faster progression of CKD in men. Females are more likely to develop CKD but have better prognosis whereas men have relatively higher mortality rate and bad prognosis. In 2017, the CKD prevalence in Central America and Mexico was estimated at 11.9% of the population.71 Prevalence was slightly higher in females (12.3%) than in males (11.5%). CKD represented 7.6% of total mortality in 2017, ranking as the second leading cause of death. The age-standardized CKD mortality rate was 42.1 per 100,000 population, a 60.9% increase between 1990 and 2017, ranging from 71.4 per 100,000 in El Salvador to 13.0 per 100,000 in Honduras. The age-standardized CKD mortality rate was higher among males.5,71 The highest estimated rates of age-standardized CKD DALYs were in El Salvador (1,821), Mexico (1,652), Nicaragua (1,571), and Guatemala (1,072), all more than 1,000 DALYs per 100,000. The age-standardized CKD DALY was higher among males (1,379) than in females (1,151). While verbal abuse does not directly cause CKD, the psychological stress and emotional distress it can cause can contribute to the development and progression of CKD by affecting blood pressure, blood vessels, and overall health. So it's just another contributing factor of mental health and impacts passively. There are other important factors that has direct impact on CKD. There's no direct impact of employment status to CKD. Rather employment status has a significant impact on physical and mental health which passively affect health and thereby affect the kidney. On the other hand, economic status has a significant socio-economic role on CKD as patients struggle with treatment and dialysis expenditure which plays a negative role over prognosis of CKD. Alcohol increases kidney filtration function and alters absorption level of medicines causing injurious impact on kidneys. On the other hand, excess alcohol increases blood pressure by means of increased vasoconstriction causing increased total peripheral resistance. A study reported that 70% of patients with cirrhosis had AKI, 17% had AKI and CKD, and 13% had CKD alone. So liver disease associated with alcohol can cause Hepatorenal Syndrome, Acute kidney injury and Metabolic Syndrome as well as oxidative stress or inflammation. Thus, it can be said that alcohol has quite a remarkable impact on CKD. Low educational attainment is associated with adverse outcomes and CKD etiology. Lifestyle habits and biomarkers mediate associations between low educational attainment and mortality. Mainly lack of education plays a pivotal role in developing CKD by means of lack of awareness, lack of health issue addressing and not taking proper health advice[118]. A high BMI was associated with poor renal outcomes in nonanemic (Hb $\geq$ 11 g/dL) patients, as well as those with CKD stages 1–3. Thus it means there is no direct relation between BMI and CKD but people with high BMI are more likely to have other comorbidities which passively impact on renal function. Undoubtedly it's one of the major causes of developing CKD. Taking painkillers, high salt intake, alcohol consumption, drinking less than required water, high sugar intake, high protein/meat intake etc directly affect the kidney system negatively and thereby contributes to CKD on a major scale[116]. Smoking can seriously harm the kidneys in a number of ways. It can increase your risk of developing some kidney cancers that damage your heart and blood vessels (cardiovascular system), leading to poor blood flow to the kidneys and causing kidney damage over time. Cardiovascular damage can also increase your risk of stroke and heart attack to help to cause or advance diabetic kidney disease – the most common cause of kidney failure. It can lead to surges in blood pressure Kidney disease and its associated complications can kill –that's why it's so important to stop smoking[117]. While physical activity is directly not related to CKD, it's connected to the functional condition of the kidney. Physical activity improves kidney function and cardiorespiratory function. Lack of physical activity or in other words sedentary lifestyle affects kidney function negatively and thereby may contribute to CKD passively. Although it may not seem connected, there's a connection at the embryological developmental stage where both the kidney and part of the ears are developed. So due to any developmental defect, both might be affected. Then again, there are some drugs, such as aminoglycosides and loop diuretics, which can cause toxicity to ears and thereby cause deafness. So, it can be said that there's some connection between deafness and CKD, which can't be overlooked, though it might not be a major concern. Though the health insurance issue is not a clinical cause, it affects the fate of CKD for most of the patients due to coverage issues. CKD needs early diagnosis and treatment for better outcomes. Patients who are unable to access healthcare facilities, including diagnosis, investigations, medications, and dialysis in need, are more prone to advancing CKD. So, from that viewpoint, affordability certainly has a socio-economic role on CKD. Although there's no significant impact of touched sexuality on CKD, CKD itself affects the sexual life/experience of the patient, resulting in sexual dysfunction. Though there's no direct impact of sexual harassment on CKD, it affects kidney function by means of psychological stress/depression-related negative impact and raises blood pressure, further contributing to the advancement of CKD. Housing instability is a significant social determinant of health that can negatively impact CKD outcomes. Addressing housing needs through screening, interventions, and advocacy is crucial for improving the health and well-being of individuals with CKD. CKD affects people of all racial and ethnic backgrounds. However, due to genetic, socioeconomic, and healthcare access barriers, Black, Hispanic or Latino, Asian, and Native American communities are at a higher risk[115].